\documentclass[11pt]{article}

\usepackage[margin=1in]{geometry}
\usepackage{amsmath,amssymb}
\usepackage{booktabs}
\usepackage{graphicx}
\usepackage{microtype}
\usepackage{hyperref}
\usepackage{xcolor}
\usepackage{authblk}
\usepackage{caption}
\usepackage{multirow}
\usepackage{tikz}
\usetikzlibrary{positioning, arrows.meta, fit, backgrounds, calc}
\usepackage[round,sort&compress,numbers]{natbib}

\hypersetup{colorlinks=true, linkcolor=blue!50!black,
            citecolor=blue!50!black, urlcolor=blue!50!black}

\title{An Empirical Audit of Input Encoders for\\
       Multi-Channel Signal Transformers}

\author{Ossi Lehtinen%
  \thanks{\texttt{ossi@ocon.fi}.
    Anthropic's Claude Code and the Opus~4.6 and 4.7 models were
    extensively used in preparing the experiments and writing this
    article.}}
\affil{Ocon Oy}
\date{}

\begin{document}
\maketitle

\begin{abstract}
Transformers consuming multi-channel scalar signals must embed $C$
simultaneous values into one $d_{\text{model}}$-dimensional vector
per time step. We audit eight input encoders---a shared-scalar
baseline, per-channel linear projections, an orthogonality
regulariser, a nonlinear MLP stem, block-partitioned concatenation,
channel-independent and channel-as-token architectures, and a
projected positional encoding---on a synthetic benchmark where
channel identity is informative and on ETTh1, scored by next-step
negative log-likelihood. The headline is practical near-equivalence
within a wide ``top tier'': the standard per-channel linear
projection matches every alternative up to small, statistically
real but practically modest differences. A direct geometric probe
attributes this to a \emph{spontaneous orthogonalisation} of the
per-channel projections: with no explicit regulariser they tighten
well beyond the near-orthogonality of random initialisation,
letting the standard linear recover channel identity from the
summed embedding. Two encoders lose decisively:
the shared-scalar baseline collapses for information-theoretic
reasons we make explicit, and the channel-independent
PatchTST-spirit baseline overfits universally on the synthetic
benchmark and underperforms on both. Paired tests resolve two
small gaps: projecting the sinusoidal positional encoding through
a learned linear layer edges the rest at small $C$ by extending
this orthogonality to the positional subspace; a nonlinear MLP
stem edges them at the largest $C$, with the gap shrinking under
more training data. The practical recommendation: use the standard
per-channel linear projection by default; reach for something
more elaborate only when the task calls for it. Code and data to reproduce every experiment in this
paper are available on GitHub.\footnote{\url{https://github.com/OssiLehtinen/channel-encoder-audit}}
\end{abstract}

\section{Introduction}

Transformers applied to multivariate numerical signals---multi-sensor
telemetry, financial factor stacks, biomedical waveforms---face a
design choice at the input layer: how to embed $C$ simultaneous scalar
channels into one $d_{\text{model}}$-dimensional vector per time step.
The central objective is to prepare the input information for
the subsequent transformer layer with minimal losses, thus allowing
maximal flexibility for modeling the influence and interplay of the
signals in relation to the target variables.

Many approaches have been put forward for achieving just this.
The additive embedding recipe inherited from NLP
\citep{vaswani2017attention} projects each input to a vector and sums.
When the inputs are multiple channels rather than one token-plus-position
pair, this comes in two forms: a \emph{shared} scalar projection $W$
with per-channel additive embeddings, or a \emph{per-channel} projection
$W_k$---the latter being mathematically a single
\texttt{nn.Linear(C, $d_{\text{model}}$)}. Many multivariate
time-series transformers use a linear input stem
\citep{zhou2021informer,wu2023timesnet}, though details vary.
Alternative architectures treat each channel as a separate token
(iTransformer \citep{liu2024itransformer}), run independent backbones
per channel (PatchTST \citep{nie2023patchtst}), or combine both
(Crossformer \citep{zhang2023crossformer}). Orthogonality penalties on
weight matrices have been used for feature decorrelation in deep
networks \citep{bansal2018gainsorthogonality}.

Despite the variety of approaches, several basic questions remain
unanswered in the literature:
(i)~Compared to a bare \texttt{nn.Linear(C, $d_{\text{model}}$)}
projection, what do we actually gain---in held-out loss, in
geometric structure, or in convergence speed---from more involved 
architectures, such as block
partitioning, an orthogonality penalty, a nonlinear MLP stem, or a
channel-independent / channel-as-token architecture?
(ii)~Does the task loss itself produce per-channel separation
without explicit enforcement?
(iii)~Does the positional encoding interact with the channel
encoding, and can this interaction be improved?
We include a shared-scalar baseline (\textsc{sum}) as a verified
floor rather than an open question---the encoder algebraically
collapses $C$ channels to their pointwise sum, so its failure mode
is information-theoretic and the only quantity of interest is how
much downstream capacity fails to recover from it.

Here, we compare eight encoders on a controlled synthetic benchmark
designed to make channel identity informative, with supporting
experiments on the public ETTh1 dataset
\citep{zhou2021informer}. Three answers map onto the three open
questions above. First, the per-channel linear projection matches
every alternative at convergence up to small, statistically real
but practically modest differences---the encoder landscape inside
the per-channel-$W_k$ family is one of practical near-equivalence
at $20$ paired seeds. Second, the task loss alone drives the
learned $W_k$ toward near-orthogonality and scales their norms
with channel importance, with no explicit regulariser required.
Third, a learned linear projection of the sinusoidal positional
encoding gives a small but consistent gain, and a direct geometric
probe shows the mechanism is positional--channel orthogonalisation
rather than positional subspace compression. The shared-scalar
\textsc{sum} baseline, included as a verified floor, hits the
capacity-independent ceiling its information-theoretic derivation
predicts. Two narrow exceptions to the practical-near-equivalence
verdict survive paired analysis but stay small in magnitude:
\textsc{mlp} edges the linear family at $C{=}16$ with the gap
shrinking ${\sim}3\times$ under $10\times$ training data, and
\textsc{cat} posts the lowest mean NLL on ETTh1 but is
statistically tied with \textsc{linear} and \textsc{linear-ortho}.

\section{Method}
\label{sec:method}

\subsection{Model architecture}

\begin{figure}[t]
\centering
\begin{tikzpicture}[
  font=\small,
  node distance=3.5mm and 10mm,
  ioblock/.style ={rectangle, draw, rounded corners=2pt,
                   fill=yellow!10, minimum width=5.2cm,
                   minimum height=0.6cm, align=center},
  encblock/.style={rectangle, draw, thick, rounded corners=2pt,
                   fill=blue!12, minimum width=5.2cm,
                   minimum height=1.0cm, align=center},
  txblock/.style ={rectangle, draw, rounded corners=2pt,
                   fill=gray!10, minimum width=5.2cm,
                   minimum height=0.6cm, align=center},
  insetblock/.style={rectangle, draw, rounded corners=1pt,
                     fill=gray!5, minimum width=4.0cm,
                     minimum height=0.45cm, align=center,
                     font=\footnotesize},
  arr/.style     ={-{Stealth[length=2.4mm]}, semithick},
  dimlabel/.style={font=\footnotesize\itshape, anchor=west, inner sep=2pt}
]
  \node[ioblock]                  (input) {Input
        $\mathbf{x} \in \mathbb{R}^{B \times T \times C}$};
  \node[encblock, below=of input] (enc)   {\textbf{\textsc{Encoder}}\\
        \emph{(the swap point: one of eight variants)}};
  \node[txblock, below=of enc]    (b1)    {Transformer block 1};
  \node[txblock, below=of b1]     (b2)    {Transformer block 2};
  \node[txblock, below=of b2]     (b3)    {Transformer block 3};
  \node[txblock, below=of b3]     (ln)    {Final LayerNorm};
  \node[txblock, below=of ln]     (head)  {Linear head
        $\mathbb{R}^{d_{\text{model}}} \to \mathbb{R}^{K}$};
  \node[ioblock, below=of head]   (out)   {Logits
        $\boldsymbol{\ell} \in \mathbb{R}^{B \times T \times K}$,
        cross-entropy on next-step bin};

  \foreach \src/\dst in {input/enc, enc/b1, b1/b2, b2/b3, b3/ln,
                         ln/head, head/out} {
    \draw[arr] (\src) -- (\dst);
  }

  \node[dimlabel] at ($(enc.south east)!0.5!(b1.north east)+(0.15,0)$)
       {$\mathbf{h} \in \mathbb{R}^{B \times T \times d_{\text{model}}}$
        \;(residual stream starts here)};

  \node[insetblock, right=14mm of b2] (z1) {LayerNorm};
  \node[insetblock, below=2mm of z1]  (z2) {Self-attention (causal)};
  \node[insetblock, below=2mm of z2]  (z3) {$+$ \emph{residual}};
  \node[insetblock, below=2mm of z3]  (z4) {LayerNorm};
  \node[insetblock, below=2mm of z4]  (z5) {FFN (GELU)};
  \node[insetblock, below=2mm of z5]  (z6) {$+$ \emph{residual}};
  \foreach \src/\dst in {z1/z2, z2/z3, z3/z4, z4/z5, z5/z6} {
    \draw[arr] (\src) -- (\dst);
  }
  \node[font=\footnotesize, above=1.5mm of z1, anchor=south]
       {\emph{Inside one block (pre-LN)}};
  \draw[dashed, gray] (b2.east) -- (z1.west);

\end{tikzpicture}
\caption{Architecture. The diagram shows the encoder-swap setup
used by six of the eight variants
(\textsc{sum}, \textsc{linear}, \textsc{linear-ortho}, \textsc{mlp},
\textsc{linear-ppe}, \textsc{concat}): the same causal transformer
backbone (three pre-LN blocks, each structured as in the inset on
the right: LN $\to$ self-attention $\to$ residual, LN $\to$ FFN
$\to$ residual; followed by a final LayerNorm and a linear head
into $K{=}32$ bin logits) is used throughout, and \textbf{the only
component that varies is the encoder (blue)}. These six encoders
share the input/output signature $\mathbb{R}^{B\times T\times C}
\to \mathbb{R}^{B\times T\times d_{\text{model}}}$ and are drop-in
interchangeable. The remaining two variants---\textsc{ci}
(channel-independent) and \textsc{cat} (channel-as-token)---are
full-architecture alternatives, not encoder swaps: \textsc{ci}
runs the transformer once per channel on $(B{\cdot}C, T,
d_{\text{model}})$ tensors and concatenates the per-channel
hiddens at the head; \textsc{cat} uses a $C{\cdot}T$-long token
sequence with attention over $(B, C{\cdot}T, d_{\text{model}})$.
Both are included as architectural baselines and are defined in
full in Section~\ref{sec:method-encoders}.}
\label{fig:arch}
\end{figure}
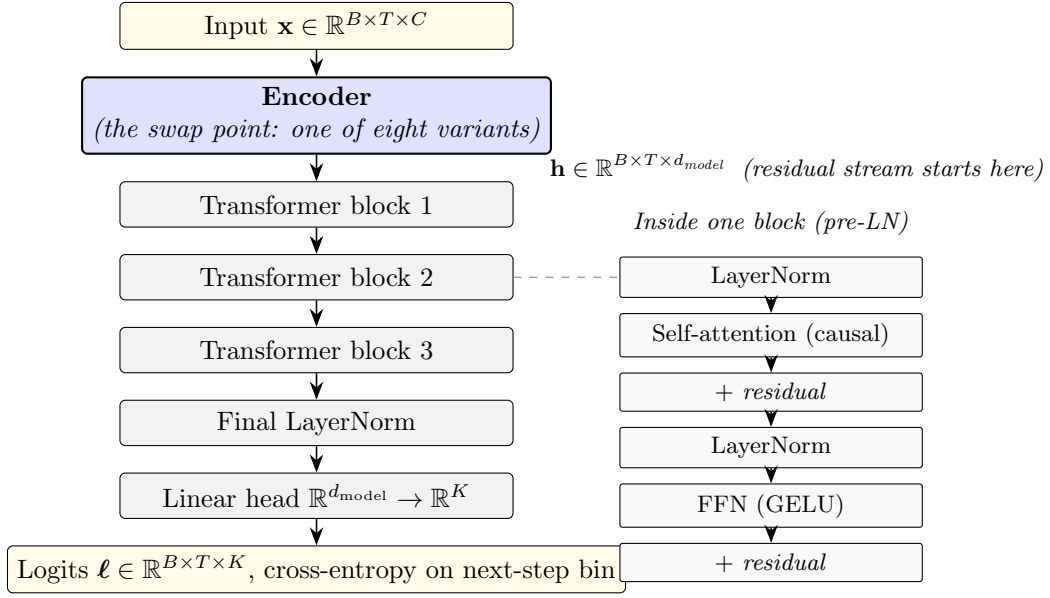

Six of the eight encoders (\textsc{sum}, \textsc{linear},
\textsc{linear-ortho}, \textsc{mlp}, \textsc{linear-ppe},
\textsc{concat}) share a common backbone: a small causal
transformer with $d_{\text{model}}{=}64$, 4 attention heads, 3
layers, feed-forward width $d_{\text{ff}}{=}256$, dropout $0.1$,
pre-LayerNorm, and a linear categorical head mapping to $K{=}32$
logits. For these six, the \emph{only} component that varies is
the input encoder (Figure~\ref{fig:arch}). The other two encoders
(\textsc{ci} and \textsc{cat}) are architectural alternatives that
reshape the token sequence the backbone consumes; their structure
is described in Section~\ref{sec:method-encoders} and we still
match $d_{\text{model}}$, head count, and layer count to the
shared backbone so the architectural cost of those alternatives is
comparable.

\subsection{Encoders}
\label{sec:method-encoders}

We study eight encoders. In the definitions below, $v_k(t)$ is the
scalar value of channel $k$ at time $t$, $\mathbf{p}(t)$ is a fixed
sinusoidal positional encoding \citep{vaswani2017attention}, and all
learned parameters are randomly initialised.

\paragraph{\textsc{sum} (shared-scalar baseline).} Shared scalar projection
$W \in \mathbb{R}^{d_{\text{model}}}$, learned per-channel additive
embedding $e_k$:
\[
  \mathbf{h}(t) = \sum_k \bigl(W\,v_k(t) + e_k\bigr) + \mathbf{p}(t).
\]
Total encoder capacity is $\mathcal{O}(d_{\text{model}})$ regardless of
$C$. We use this to isolate what goes wrong when the input projection
shares weights across channels.

\paragraph{\textsc{linear}.} Per-channel projection $W_k$ and bias $b_k$,
summed:
\[
  \mathbf{h}(t) = \sum_k \bigl(W_k\,v_k(t) + b_k\bigr) + \mathbf{p}(t).
\]
Equivalently $\mathbf{h}(t) = W\,\mathbf{v}(t) + \mathbf{b} +
\mathbf{p}(t)$ with $W \in \mathbb{R}^{d_{\text{model}}\times C}$:
a single \texttt{nn.Linear(C, $d_{\text{model}}$)}.

\paragraph{\textsc{linear-ortho}.} Same as \textsc{linear} with an
auxiliary loss $L_{\text{ortho}} = \lambda \sum_{i \ne j}
(W_i \cdot W_j)^2 / 2$ ($\lambda{=}10^{-2}$). This variant isolates the
regulariser's contribution.

\paragraph{\textsc{linear-ppe} (projected positional encoding).}
Same channel encoding as \textsc{linear}, but the sinusoidal position
passes through a learned linear layer:
\[
  \mathbf{h}(t) = \sum_k \bigl(W_k\,v_k(t) + b_k\bigr)
                + W_{\text{pos}}\,\mathbf{p}(t) + \mathbf{b}_{\text{pos}},
\]
with $W_{\text{pos}} \in \mathbb{R}^{d_{\text{model}} \times
d_{\text{model}}}$. The sinusoidal basis is otherwise fixed at
initialisation and cannot move; the learned $W_{\text{pos}}$ adds
the degree of freedom needed for the positional subspace to
relocate. The cross-stream gradient pressure already present in
\textsc{linear} (Section~\ref{sec:results-pospro} makes it
explicit) can then act on the positional side as well; what we
measure is whether the optimiser uses the new freedom to rotate
$\mathbf{p}(t)$ out of the channel subspace. Adds
$d_{\text{model}}^2 + d_{\text{model}}$ parameters (4{,}160 at
$d{=}64$).
The construction is closely related to the untied positional
embedding of \citet{ke2021rethinking}, which removes the additive
position--content coupling at the input layer of language
transformers via a separate projection; \textsc{linear-ppe} applies
the same decoupling principle to the channel--position competition
in time-series transformers.

\paragraph{\textsc{mlp}.} Two-layer MLP on the channel vector:
$\mathbf{h}(t) = W_2\,\text{GELU}(W_1\,\mathbf{v}(t) + \mathbf{b}_1)
+ \mathbf{b}_2 + \mathbf{p}(t)$, with
$W_1 \in \mathbb{R}^{d_{\text{model}} \times C}$,
$W_2 \in \mathbb{R}^{d_{\text{model}} \times d_{\text{model}}}$.
Tests whether nonlinearity helps (${\sim}9{\times}$ more encoder
parameters than \textsc{linear}).

\paragraph{\textsc{concat}.} Per-channel projection into
$d_{\text{block}} {=} d_{\text{model}}/C$ dims, concatenated
(requires $C \mid d_{\text{model}}$):
$\mathbf{h}(t) = [\mathbf{e}_0(t) \Vert \cdots \Vert
\mathbf{e}_{C-1}(t)] + \mathbf{p}(t)$. Channel identity is an
architectural invariant.

\paragraph{\textsc{ci} (channel-independent).} PatchTST-spirit
\citep{nie2023patchtst}: shared-weight causal transformer per channel;
final hidden states concatenated into the head. The computational cost
 of this approach is high with $\mathcal{O}(C\,B\,T^2)$ scaling.

\paragraph{\textsc{cat} (channel-as-token).} iTransformer-spirit
\citep{liu2024itransformer}: each $(t, k)$ pair is a token, sequence
length $C{\cdot}T$, causal across time, bidirectional within a time
step. Attention scales as $\mathcal{O}((CT)^2)$, which makes
\textsc{cat} substantially heavier than the other encoders at every
$C$ and prohibitively expensive at $C{=}16, T{=}160$ on our compute
budget; we skip $C{=}16$ for that reason.

\subsection{Training protocol}
\label{sec:method-training}

Identical across encoders and across both datasets unless noted.

\paragraph{Objective.} At every position $t \in \{0, \dots, T{-}1\}$
the model emits logits $\ell_t \in \mathbb{R}^{K}$ and we minimise the
next-step categorical cross-entropy (negative log-likelihood, NLL)
\[
\mathcal{L}_{\text{NLL}}
  = -\frac{1}{B(T{-}1)} \sum_{b=1}^{B} \sum_{t=0}^{T-2}
      \log\,\mathrm{softmax}(\ell_t^{(b)})_{\,y^{(b)}_{t+1}},
\]
where $y^{(b)}_t$ is the bin index at position $t$ in series $b$. The
predicted distribution at position $t$ is supervised only by the
bin at position $t+1$, and causal masking inside the transformer
prevents the model from seeing $y_{t+1}$ at prediction time.
\textsc{linear-ortho} adds the auxiliary orthogonality term
defined in Section~\ref{sec:method-encoders}; all other encoders
have $\mathcal{L} = \mathcal{L}_{\text{NLL}}$. As a
loss-family robustness check we also re-run a subset of the
synthetic sweep with a scalar regression head and MSE loss on the
continuous (pre-binning) target; this is described in
Section~\ref{sec:results-mse}.

\paragraph{Optimiser.} AdamW with peak learning rate $3{\times}10^{-4}$,
weight decay $10^{-4}$, and PyTorch defaults for the moment
estimates ($\beta_1{=}0.9$, $\beta_2{=}0.999$,
$\epsilon{=}10^{-8}$). Gradient norm is clipped to $1.0$ per step,
batch size $32$. The learning rate follows a cosine annealing
schedule from peak to $3{\times}10^{-6}$ ($1\%$ of peak) over the
full $300$ epochs, with no warm-up: the pre-LN architecture
\citep{xiong2020layernorm} is gradient-stable at initialisation
and our convergence flags across the $20$-seed sweep
(Section~\ref{sec:results-convergence}) show no early instability
on any of the top-tier encoders.

\paragraph{Train/val split.}
For the synthetic benchmark we generate $n_{\text{series}}{=}512$
length-$T{=}160$ series per seed and split with a seeded
\texttt{torch.utils.data.random\_split} into
$461 / 51$ ($\approx 90/10$) train and val. A separate
``data-rich'' condition with $n_{\text{series}}{=}5120$
($10{\times}$ training data, same $90/10$ split) is used at $C{=}16$
to check whether the encoder ordering at our main-sweep scale is
data-limited; this condition is reported in
Section~\ref{sec:results-scaling}.
For ETTh1 we follow the dataset's standard chronological partition,
using the train block to fit per-variate standardisation statistics
and the quantile bin edges for the target, and evaluating on the
val block.

\paragraph{Validation cadence and model selection.} Validation
NLL and top-$1$ accuracy are evaluated every $20$ epochs (plus
epochs $1$ and $300$). We report each run's best validation NLL
along this trace and the accuracy at the same epoch---``effective
early stopping'', applied uniformly. Training is never halted;
every run sees the full $300$-epoch schedule.

\paragraph{Seeds and reproducibility.}
Headline configurations are run over $20$ random seeds, indexed $0$
through $19$. Seed $s$ sets \texttt{torch.manual\_seed(s)}
(controlling weight initialisation and dropout) and seeds the
\texttt{torch.Generator} passed to \texttt{random\_split}
(controlling the train/val partition); on the synthetic benchmark
it also drives the \texttt{numpy} RNG that generates the input
series. \emph{The same $20$ seeds are reused for every encoder
variant, every $C$, and every $d_{\text{model}}$.} This is a
paired-seed design: at a fixed $s$, every encoder sees identical
training data and the identical train/val partition, and only the
model architecture (and its initial parameter draws from the shared
PRNG state) differs. We \emph{report} unpaired mean$\,\pm\,$std
intervals throughout for readability, but we \emph{test} with
paired-difference statistics where the unpaired interval is loose
enough that paired analysis changes the conclusion. Where the
paired and unpaired test agree qualitatively (most rows in this
paper) we report only the unpaired interval; where they disagree,
or where the unpaired $p$-value is borderline, we report the paired
result and call it out. A few diagnostics that consume substantially
more compute or are by-design single-instance---linear probing,
test-time channel masking, the channel-bias ablation, and the
\textsc{mlp} first-layer geometry table---are kept at $5$ seeds and
labelled as such in their captions. Bootstrap confidence intervals
on cosine, paired-difference, and similar derived quantities are
constructed by resampling seeds (and, where applicable, examples)
with $10{,}000$ resamples; reported as $95\%$ intervals. We do not
enable \texttt{torch.use\_deterministic\_algorithms(True)}, so CUDA
matmul nondeterminism contributes some additional seed-level
variation between full re-runs; the reported $\pm\,$std intervals
capture this.

\subsection{Datasets}

\paragraph{Synthetic.} $N{=}512$ time series of length $T{=}160$ with
$C$ channels. Each channel is a sum of three sinusoids with random
frequencies in $[0.005, 0.08]$ cycles/sample plus AR(1) noise,
standardised. The outcome is a fixed non-linear function of the first
four channels with lags 3 and 7:
\[
y_t = \tanh\!\big(s_0(t{-}3)\,s_1(t)\big)
    + 0.6\sin\!\big(1.3\,s_2(t{-}7)\big)
    + 0.4\,\mathbb{1}[s_3(t)>0]\,s_0(t),
\]
binned into $K{=}32$ quantile bins. Channels $k \ge 4$ are independent
distractors. The task is designed so that an encoder that mixes channels
at the input layer cannot recover the interaction structure.

\paragraph{ETTh1.} Electricity Transformer Temperature
\citep{zhou2021informer}, hourly, 7 variates. We standardise on the
train split, bin the target variate
OT into 32 quantile bins, and predict the next-step bin given all 7
variates over $T{=}160$. The model uses $d_{\text{model}}{=}56$
(divisible by $C{=}7$), 7 heads, $d_{\text{ff}}{=}224$; training is
otherwise identical.

\subsection{Diagnostic analyses}
\label{sec:method-diagnostics}

Beyond aggregate NLL, we use three diagnostics to inspect what the
per-channel-$W_k$ encoders actually learn. Each is computed after
full 300-epoch training and aggregated over $20$ seeds unless
stated otherwise (the channel-bias ablation, linear probing,
test-time channel masking, and the \textsc{mlp} first-layer
geometry table use $5$ seeds).

\paragraph{Gram-matrix analysis.}
\textsc{linear} and \textsc{linear-ortho} differ only in the
auxiliary orthogonality penalty (Section~\ref{sec:method-encoders}).
If both reach the same NLL, the question is whether
\textsc{linear}'s $W_k$ are geometrically similar to
\textsc{linear-ortho}'s---i.e.\ whether the task pressure alone
already produces a near-orthogonal arrangement---or whether the two
encoders arrive at the same loss through different geometries.

After training we extract the per-channel projection matrix $W \in
\mathbb{R}^{C \times d_{\text{model}}}$ (rows $W_k$) from the encoder
and compute the cosine matrix $G_{ij} = (W_i \cdot W_j) / (\|W_i\|
\|W_j\|)$. Two scalar summaries:
\[
\overline{|\cos|} = \frac{1}{C(C-1)} \sum_{i \ne j} |G_{ij}|,
\qquad
\max|\cos| = \max_{i \ne j} |G_{ij}|.
\]
Both go to zero iff the $W_k$ become mutually orthogonal. We report
seed-averaged means. Because random high-dimensional vectors are
already nearly orthogonal (i.i.d.\ Gaussian rows in
$d_{\text{model}}$ dimensions have expected off-diagonal
$|\cos| \approx \sqrt{2/(\pi d_{\text{model}})}$), we also compute
both summaries at initialisation by re-instantiating each seed's
initial parameter draw, so that trained values can be compared
seed-paired against the baseline that high-dimensional geometry
provides for free.

\paragraph{Encoding-space allocation.}
Orthogonality says channels are distinguishable; it does not say
whether the model dedicates more representational capacity to
outcome-relevant channels than to distractors. We measure two
quantities, both at the encoder output:

(i)~Norm $\|W_k\|$ for each channel. A larger norm means channel $k$'s
contribution $W_k v_k(t) + b_k$ has larger magnitude in the embedding,
i.e.\ occupies more of the residual stream.

(ii)~Per-channel variance fraction. The encoder output is a sum of
per-channel contributions $W_k v_k(t) + b_k$. Treating each
contribution as a random vector indexed by $(t, \text{batch})$, the
per-channel variance is
\[
\sigma^2_k = \sum_{d=1}^{d_{\text{model}}}
             \operatorname{Var}_{t,\text{batch}}\!\big[(W_k v_k(t) + b_k)_d\big],
\]
and we report fractions $\sigma^2_k / \sum_j \sigma^2_j$ on a held-out
batch. This combines the size of $W_k$ with the empirical variability
of channel $k$'s input distribution.

For both metrics, channels are partitioned into drivers ($k \in
\{0,1,2,3\}$, which enter the outcome formula) and distractors
($k \ge 4$, independent of the outcome). Reporting them this way lets
us check whether the model independently discovered the
driver/distractor split.

\paragraph{Linear-probing analysis.}
The encoder lays down a representation; the question is whether
downstream attention and FFN layers preserve enough of it that
each channel remains linearly identifiable several blocks deep.
The pre-LayerNorm transformer's residual structure already makes
a zeroth-order prediction here: each block applies $h \leftarrow
h + \text{Attn}(\text{LN}(h))$ and $h \leftarrow h +
\text{FFN}(\text{LN}(h))$, so an unmodified copy of the input
embedding is carried forward through every block. Under the
residual-stream view formalised by \citet{elhage2021framework},
the residual stream is a communication bus that sublayers read
from and additively write into; we should therefore expect deep
linear probes \citep{alain2017probes} to recover whatever was
written to the residual stream at layer~$0$, modulo what LayerNorm
rescales and what the sublayers additively perturb. We test this
directly by training closed-form ridge probes from a frozen hidden
state back to the raw input.

Procedure: (1)~train the encoder--transformer end-to-end as usual;
(2)~freeze it; (3)~generate a fresh probe dataset
(\texttt{MultiSignalDataset} with a different seed, so the model has
not been trained on it); (4)~for each layer of interest, pass the
probe dataset through and collect hidden states
$H \in \mathbb{R}^{N \times d_{\text{model}}}$ alongside the
corresponding raw inputs $X \in \mathbb{R}^{N \times C}$; (5)~solve
the ridge regression
\[
W_{\text{probe}} = (H^\top H + \lambda I)^{-1} H^\top X,
\qquad \lambda = 10^{-3},
\]
in closed form on a probe-train split; (6)~evaluate held-out
$R^2$ per channel on a probe-validation split:
$R^2_k = 1 - \mathrm{SS}_{\text{res},k} / \mathrm{SS}_{\text{tot},k}$.

We probe two layers: layer 0 (the encoder output, i.e.\ the input to
the first transformer block) and layer 3 (the output of the third and
final transformer block, before the LayerNorm and the categorical
head). $R^2 = 1$ at layer 0 confirms the encoder writes the raw
channel into a linearly recoverable subspace; $R^2$ at layer 3 reveals
how much of that recoverability survives the depth of the model.
\textsc{ci} and \textsc{cat} are excluded because their hidden states
have a fundamentally different shape (per-channel stream and per-token
respectively), so a single $H \to X$ probe does not apply.

\section{Results}

\subsection{Main comparison}
\label{sec:results-main}

\begin{table}[t]
\centering
\caption{Main comparison at $C{=}4$, $d_{\text{model}}{=}64$, 300 epochs,
cosine LR decay, mean $\pm$ std over 20 seeds (5 canonical + 15
paired-seed extras). Best val NLL (effective early stopping). Random
baseline: NLL$\,{=}\ln 32 \approx 3.47$, acc$\,{=}0.031$.
$^\dagger$For \textsc{ci} and \textsc{cat}, ``encoder params'' counts
only the per-channel input projection (plus per-variate embedding for
\textsc{cat}); the bulk of those architectures' parameters lives in
the replacement backbone.}
\label{tab:main}
\begin{tabular}{lccc}
\toprule
Encoder & Enc.\ params & Val NLL $\downarrow$ & Val acc $\uparrow$ \\
\midrule
\textsc{sum}        & 320     & $3.257 \pm 0.011$ & $0.091 \pm 0.003$ \\
\textsc{ci}$^\dagger$ & 128   & $3.053 \pm 0.013$ & $0.136 \pm 0.004$ \\
\textsc{cat}$^\dagger$ & 384  & $2.348 \pm 0.060$ & $0.212 \pm 0.011$ \\
\midrule
\textsc{concat}     & 128     & $2.170 \pm 0.025$ & $0.243 \pm 0.006$ \\
\textsc{linear}     & 512     & $2.155 \pm 0.019$ & $0.247 \pm 0.006$ \\
\textsc{linear-ortho}  & 512     & $2.155 \pm 0.020$ & $0.247 \pm 0.006$ \\
\textsc{mlp}        & 4{,}480 & $2.177 \pm 0.022$ & $0.242 \pm 0.007$ \\
\textsc{linear-ppe} & 4{,}672 & $2.114 \pm 0.029$ & $0.256 \pm 0.009$ \\
\bottomrule
\end{tabular}
\end{table}

Table~\ref{tab:main} reports all eight encoders at $C{=}4$,
$d_{\text{model}}{=}64$, 300 epochs with cosine LR decay, 20 seeds.
The headline read is \emph{practical near-equivalence among the
per-channel-$W_k$ family}: \textsc{linear}, \textsc{linear-ortho},
\textsc{concat}, and \textsc{mlp} all fall in a $0.02$ NLL band at
$2.155$--$2.177$, with \textsc{linear} and \textsc{linear-ortho}
matching to three decimals (the orthogonality penalty contributes
nothing at convergence). \textsc{linear-ppe} sits a small step
below at $2.114$, a ${\sim}0.04$ NLL improvement over
\textsc{linear} that survives paired analysis at $20$ seeds; the
mechanism is a learned rotation of the positional encoding out of
the channel subspace, established in
Section~\ref{sec:results-pospro}. The gap corresponds to about
$2\%$ of the baseline NLL and is well within the spread that
small task-design choices would induce; we read it as a real
effect, not a practically transformative one. Below the top tier, the architectural alternatives underperform
despite nominally giving each channel more capacity. \textsc{cat}
sits ${\sim}0.18$ NLL below the per-channel-$W_k$ family at
$2.348$. \textsc{ci} drops decisively further at $3.053$, only
${\sim}0.2$ NLL above the \textsc{sum} floor of $3.257$ and almost
$0.9$ NLL below the top tier; it is closer to the shared-scalar
information-theoretic floor than to the linear baseline. So two
encoders lose decisively on the synthetic benchmark:
\textsc{sum}, for the algebraic reasons discussed in the
Discussion, and \textsc{ci}, whose head-side bottleneck and
universal overfitting (every seed flags as overfitting) place
it close to the \textsc{sum} floor on this benchmark.

\subsection{Channel-count scaling}
\label{sec:results-scaling}

\begin{table}[t]
\centering
\caption{Scaling with channel count. Channels $0$--$3$ drive the
outcome; additional channels are independent distractors. 300 epochs
with cosine decay, mean $\pm$ std over 20 seeds. \textsc{cat} is
skipped at $C{=}16$ because its $\mathcal{O}((CT)^2)$ attention cost
explodes. Sub-band sub-tables span the within-cell standard error
times a t-quantile; encoder-vs-encoder paired tests reported in the
text use the same paired-seed design across encoders.}
\label{tab:scaling}
\begin{tabular}{llcc}
\toprule
$C$ & encoder & val NLL $\downarrow$ & val acc $\uparrow$ \\
\midrule
\multirow{8}{*}{4}
    & \textsc{sum}        & $3.257 \pm 0.011$ & $0.091 \pm 0.003$ \\
    & \textsc{ci}         & $3.053 \pm 0.013$ & $0.136 \pm 0.004$ \\
    & \textsc{cat}        & $2.348 \pm 0.060$ & $0.212 \pm 0.011$ \\
    & \textsc{concat}     & $2.170 \pm 0.025$ & $0.243 \pm 0.006$ \\
    & \textsc{mlp}        & $2.177 \pm 0.022$ & $0.242 \pm 0.007$ \\
    & \textsc{linear-ortho}  & $2.155 \pm 0.020$ & $0.247 \pm 0.006$ \\
    & \textsc{linear}     & $2.155 \pm 0.019$ & $0.247 \pm 0.006$ \\
    & \textsc{linear-ppe} & $2.114 \pm 0.029$ & $0.256 \pm 0.009$ \\
\midrule
\multirow{8}{*}{8}
    & \textsc{sum}        & $3.300 \pm 0.006$ & $0.081 \pm 0.004$ \\
    & \textsc{ci}         & $3.062 \pm 0.016$ & $0.135 \pm 0.004$ \\
    & \textsc{cat}        & $2.334 \pm 0.073$ & $0.213 \pm 0.011$ \\
    & \textsc{concat}     & $2.234 \pm 0.023$ & $0.231 \pm 0.005$ \\
    & \textsc{mlp}        & $2.211 \pm 0.027$ & $0.236 \pm 0.005$ \\
    & \textsc{linear-ortho}  & $2.176 \pm 0.022$ & $0.244 \pm 0.006$ \\
    & \textsc{linear}     & $2.176 \pm 0.022$ & $0.243 \pm 0.006$ \\
    & \textsc{linear-ppe} & $2.150 \pm 0.030$ & $0.249 \pm 0.006$ \\
\midrule
\multirow{7}{*}{16}
    & \textsc{sum}        & $3.310 \pm 0.004$ & $0.075 \pm 0.003$ \\
    & \textsc{ci}         & $3.078 \pm 0.017$ & $0.133 \pm 0.005$ \\
    & \textsc{concat}     & $2.340 \pm 0.033$ & $0.215 \pm 0.008$ \\
    & \textsc{linear-ortho}  & $2.267 \pm 0.038$ & $0.227 \pm 0.007$ \\
    & \textsc{linear}     & $2.267 \pm 0.032$ & $0.226 \pm 0.007$ \\
    & \textsc{linear-ppe} & $2.252 \pm 0.038$ & $0.230 \pm 0.008$ \\
    & \textsc{mlp}        & $2.231 \pm 0.032$ & $0.234 \pm 0.004$ \\
\bottomrule
\end{tabular}
\end{table}

\begin{figure}[t]
\centering
\includegraphics[width=0.92\textwidth]{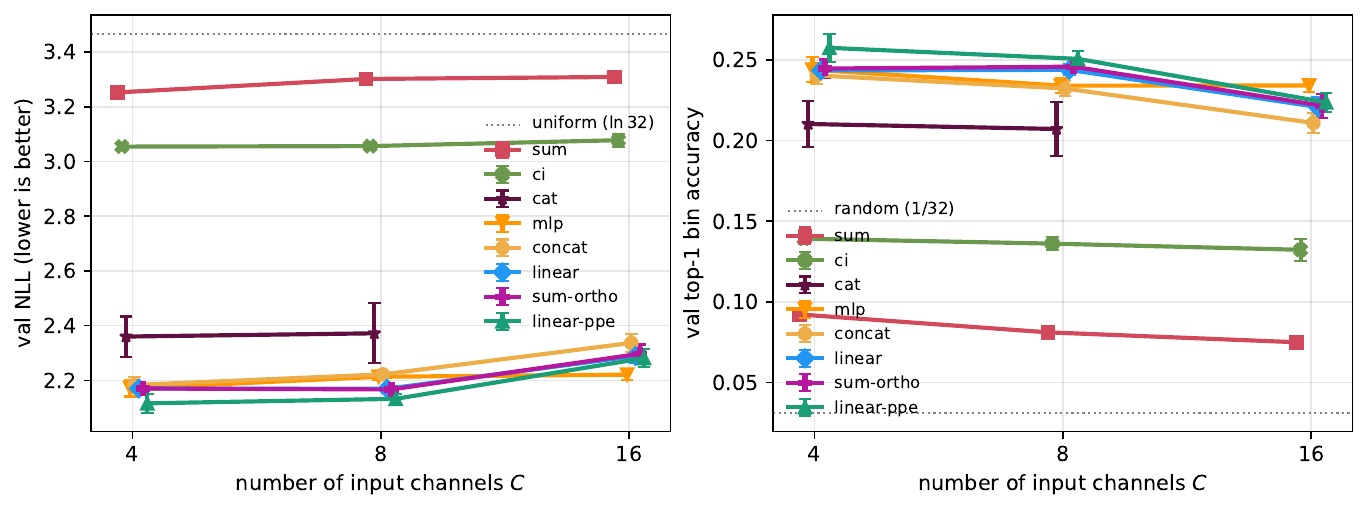}
\caption{Scaling with input channel count $C$ (error bars: $\pm 1$
standard deviation across $20$ seeds). \textsc{cat} skipped at
$C{=}16$ due to $\mathcal{O}((CT)^2)$ attention cost.}
\label{fig:scaling}
\end{figure}

All encoders degrade as $C$ grows (channels $k \ge 4$ are
distractors). The per-channel-$W_k$ family stays clustered at the
top across $C \in \{4, 8, 16\}$ and the gap to \textsc{sum} widens
rather than narrows. \textsc{linear-ppe} sits at the very top at
$C{=}4$ and $C{=}8$ and ties \textsc{mlp} at $C{=}16$ (with
\textsc{mlp} taking the lowest mean at that $C$); the paired-test
statistics, and the geometric mechanism explaining why the
\textsc{linear-ppe} edge shrinks with $C$, are in
Section~\ref{sec:results-pospro}.

At $C{=}16$, \textsc{mlp} pulls ahead of the linear family
($2.231$ vs.\ $2.27$--$2.34$). Paired tests against \textsc{linear}
($p{=}2{\times}10^{-4}$, 18/20), \textsc{linear-ortho}
($p{=}7{\times}10^{-4}$, 17/20), and \textsc{concat}
($p{=}5{\times}10^{-12}$, 20/20) are decisive; against
\textsc{linear-ppe} the gap is small and the call is borderline
($\Delta = 0.020$, paired $p{=}0.053$, 12/20). Combined with the
\textsc{linear-ppe} lead at smaller $C$, the picture is a narrow top
tier whose member ordering shifts with $C$: \textsc{linear-ppe} edges
the linear family at every $C$, \textsc{mlp} edges the linear family
(but ties \textsc{linear-ppe}) at $C{=}16$, and at every $C$ the
absolute spread among the top four is on the order of $0.02$--$0.05$
NLL on a baseline of $\sim$$2.2$ NLL.

\subsection{Data-richness check: $C{=}16$ at $10\times$ training data}
\label{sec:results-largen}

The $\Delta$NLL ${\sim}0.07$ between \textsc{mlp} and \textsc{linear}
at $C{=}16, N_{\text{series}}{=}512$ raises an obvious question: is
the nonlinear stem genuinely useful, or is the main sweep
data-limited and \textsc{mlp} just happens to extract more from
sparse data? We test this directly by retraining the same top-tier
encoders at $C{=}16$ with $N_{\text{series}}{=}5120$ ($10\times$
the main-sweep data), 20 seeds, otherwise identical.

\begin{table}[h]
\centering
\caption{C=16 top-tier encoders, $N_{\text{series}}{=}5120$ vs.\
$N_{\text{series}}{=}512$ (20 seeds each, 300 epochs). The full
linear family at $N{=}5120$ falls inside a $0.024$ NLL band.}
\label{tab:main-largen}
\begin{tabular}{lccc}
\toprule
encoder & $N{=}512$ val NLL & $N{=}5120$ val NLL & $\Delta$ from data \\
\midrule
\textsc{mlp}        & $2.231 \pm 0.032$ & $1.831 \pm 0.009$ & $-0.400$ \\
\textsc{linear-ppe} & $2.252 \pm 0.038$ & $1.841 \pm 0.013$ & $-0.411$ \\
\textsc{linear}     & $2.267 \pm 0.032$ & $1.843 \pm 0.015$ & $-0.424$ \\
\textsc{linear-ortho}  & $2.267 \pm 0.038$ & $1.844 \pm 0.018$ & $-0.423$ \\
\textsc{concat}     & $2.340 \pm 0.033$ & $1.854 \pm 0.011$ & $-0.486$ \\
\bottomrule
\end{tabular}
\end{table}

Two observations follow. \textsc{mlp}'s lead survives in the
data-rich regime: $18$--$20$ of $20$ paired-seed differences favour
\textsc{mlp} over each of \textsc{linear}
(paired $p{=}4{\times}10^{-4}$), \textsc{linear-ortho}
($p{=}9.6{\times}10^{-4}$), \textsc{linear-ppe}
($p{=}2.4{\times}10^{-4}$), and \textsc{concat}
($p{=}2.3{\times}10^{-13}$), so the $C{=}16$ \textsc{mlp} finding
is not a data-limited artefact. But the magnitude shrinks roughly
three-fold: $\Delta$NLL(\textsc{mlp}--\textsc{linear}) goes from
$0.036$ at $N{=}512$ to $0.011$ at $N{=}5120$; the full linear
family at $N{=}5120$ falls inside a band of width $0.024$ NLL
versus $0.073$ at $N{=}512$. Each encoder also gains ${\sim}0.4$ NLL from the extra
data, confirming the $N{=}512$ sweep is data-limited at $C{=}16$ just
as at $C{=}4$ (Section~\ref{sec:results-space}). The encoder ordering
at $C{=}16$ is robust to data scale; the absolute size of the gaps
is not.

\subsection{Model-width scaling}
\label{sec:results-dmodel}

\begin{table}[t]
\centering
\caption{Scaling with $d_{\text{model}}$ at $C{=}4$.
$d_{\text{ff}}{=}4\,d_{\text{model}}$, $4$ attention heads (so
$d_{\text{head}}$ grows). Mean $\pm$ std over $20$ seeds, $300$ epochs
with cosine decay. \emph{All} encoders flag as overfitting at
$d_{\text{model}} \ge 128$: best val NLL is reached well before epoch
$300$ (typical best epochs $\sim$$100$ at $d{=}128$, $\sim$$40$ at
$d{=}256$), and val NLL drifts upward afterwards. The reported
best-NLL numbers are the effective-early-stopping minima.}
\label{tab:dmodel}
\begin{tabular}{llccc}
\toprule
$d_{\text{model}}$ & encoder & total params & val NLL $\downarrow$ & val acc $\uparrow$ \\
\midrule
\multirow{6}{*}{64}
    & \textsc{sum}        & $152{,}480$    & $3.257 \pm 0.011$ & $0.091 \pm 0.003$ \\
    & \textsc{concat}     & $152{,}288$    & $2.170 \pm 0.026$ & $0.243 \pm 0.006$ \\
    & \textsc{linear}     & $152{,}672$    & $2.155 \pm 0.019$ & $0.248 \pm 0.005$ \\
    & \textsc{linear-ortho}  & $152{,}672$    & $2.155 \pm 0.019$ & $0.248 \pm 0.005$ \\
    & \textsc{mlp}        & $156{,}640$    & $2.177 \pm 0.023$ & $0.241 \pm 0.008$ \\
    & \textsc{linear-ppe} & $156{,}832$    & $2.115 \pm 0.029$ & $0.256 \pm 0.009$ \\
\midrule
\multirow{6}{*}{128}
    & \textsc{sum}        & $599{,}840$    & $3.260 \pm 0.010$ & $0.091 \pm 0.003$ \\
    & \textsc{concat}     & $599{,}456$    & $2.229 \pm 0.031$ & $0.235 \pm 0.006$ \\
    & \textsc{linear}     & $600{,}224$    & $2.234 \pm 0.032$ & $0.235 \pm 0.006$ \\
    & \textsc{linear-ortho}  & $600{,}224$    & $2.233 \pm 0.032$ & $0.235 \pm 0.007$ \\
    & \textsc{mlp}        & $616{,}352$    & $2.301 \pm 0.027$ & $0.222 \pm 0.007$ \\
    & \textsc{linear-ppe} & $616{,}736$    & $2.185 \pm 0.032$ & $0.245 \pm 0.005$ \\
\midrule
\multirow{6}{*}{256}
    & \textsc{sum}        & $2{,}379{,}296$ & $3.262 \pm 0.011$ & $0.091 \pm 0.003$ \\
    & \textsc{concat}     & $2{,}378{,}528$ & $2.311 \pm 0.026$ & $0.217 \pm 0.004$ \\
    & \textsc{linear}     & $2{,}380{,}064$ & $2.322 \pm 0.033$ & $0.217 \pm 0.006$ \\
    & \textsc{linear-ortho}  & $2{,}380{,}064$ & $2.323 \pm 0.036$ & $0.217 \pm 0.006$ \\
    & \textsc{mlp}        & $2{,}445{,}088$ & $2.407 \pm 0.023$ & $0.210 \pm 0.007$ \\
    & \textsc{linear-ppe} & $2{,}445{,}856$ & $2.265 \pm 0.026$ & $0.229 \pm 0.006$ \\
\bottomrule
\end{tabular}
\end{table}

Table~\ref{tab:dmodel} varies $d_{\text{model}}$ over $\{64, 128, 256\}$
at $C{=}4$, $20$ seeds, all top-tier encoders plus the \textsc{sum}
and \textsc{concat} baselines. Three observations.

First, \textsc{sum}'s ceiling is structural and
capacity-independent: $15{\times}$ more parameters move NLL by
less than $0.01$. The encoder destroys $(C{-}1)$ channels' worth
of information through its shared scalar projection, and
downstream capacity cannot recover what the encoder threw away
(formal derivation in the Discussion).

Second, \emph{every} encoder shows clear overfitting at
$d_{\text{model}} \ge 128$. Best val NLL is reached at epoch
$\sim$$100$ ($d{=}128$) or $\sim$$40$ ($d{=}256$), after which val
NLL drifts upward (catastrophically for \textsc{mlp} at $d{=}256$,
where the end-of-training val NLL exceeds $\ln 32$). Effective early
stopping (Section~\ref{sec:method-training}) is what protects the
reported best-NLL numbers; the trajectories themselves are not
converged. $C{=}4$ with $N_{\text{series}}{=}512$ does not justify a
$256$-dim residual stream.

Third, the linear family's best-val ceiling rises gently with
$d_{\text{model}}$ ($2.16 \to 2.23 \to 2.32$ for \textsc{linear}),
indicating that the optimal effective-stopping model gets worse as
the model is given more capacity to overfit. \textsc{mlp} degrades
fastest of the top tier ($2.18 \to 2.30 \to 2.41$) because its
encoder parameters scale as $d_{\text{model}}^2$ and exacerbate the
capacity mismatch. The paired tests against \textsc{linear} sharpen
with width: $p{=}2{\times}10^{-4}$ at $d{=}64$ (18/20 favour
\textsc{linear}), $p{=}4{\times}10^{-9}$ at $d{=}128$ (20/20), and
$p{=}5{\times}10^{-9}$ at $d{=}256$ (19/20). The $C{=}4$
small-channel regime does not reward nonlinearity, and gives it room
to overfit when widened. This is in sharp contrast to $C{=}16$ where
\textsc{mlp} leads (Section~\ref{sec:results-scaling}): the
high-channel-pressure regime is what makes nonlinearity useful.

\textsc{linear-ppe} also leads at every $d_{\text{model}}$
tested, with paired gaps to \textsc{linear} growing mildly with
width and 56 of 60 paired-seed comparisons across the three
widths favouring it. Details and the rotation-based mechanism
that explains why more residual-stream width helps are in
Section~\ref{sec:results-pospro}.

\subsection{Gram-matrix analysis: spontaneous orthogonality}
\label{sec:results-gram}

Applying the off-diagonal cosine diagnostic from
Section~\ref{sec:method-diagnostics} to the trained $W_k$:

\begin{table}[h]
\centering
\caption{Per-channel $W_k$ geometry ($C{=}4$, $300$ epochs,
mean $\pm$ std over $20$ seeds; same runs as Table~\ref{tab:main}).
The initialisation row re-instantiates the same $20$ per-seed
initial parameter draws used by those runs.}
\label{tab:gram}
\begin{tabular}{lccc}
\toprule
encoder      & best val NLL & $\max_{i\ne j}|\cos(W_i,W_j)|$ & mean $|\cos|$ \\
\midrule
at initialisation    & --- & $0.219 \pm 0.061$ & $0.109 \pm 0.032$ \\
\textsc{linear}      & $2.155 \pm 0.019$ & $0.095 \pm 0.032$ & $0.042 \pm 0.010$ \\
\textsc{linear-ortho} & $2.155 \pm 0.020$ & $0.022 \pm 0.008$ & $0.010 \pm 0.003$ \\
\bottomrule
\end{tabular}
\end{table}

At $d_{\text{model}}{=}64$ the initialisation is itself
near-orthogonal, as high-dimensional geometry predicts (measured
mean $|\cos|$ $0.109$ against the analytic expectation
${\approx}\,0.10$), so a near-orthogonal trained geometry would by
itself demonstrate nothing. The comparison that matters is paired
against each seed's own initial draw: training tightens the mean
off-diagonal cosine from $0.109$ to $0.042$ (${\sim}2.6\times$;
paired per-seed difference $-0.068$, $95\%$ bootstrap CI
$[-0.080, -0.054]$, tighter in $20/20$ seeds) and the maximum from
$0.219$ to $0.095$ (paired difference $-0.124$, CI
$[-0.154, -0.095]$, tighter in $19/20$ seeds). The regulariser
tightens the mean by another factor of ${\sim}4$ to $0.010$
($95\%$ CI $[0.008, 0.011]$) but does not improve downstream NLL.
The task loss actively orthogonalises the per-channel projections
well beyond the random-initialisation default; the regulariser
only sharpens a geometry that emerges on its own.

\paragraph{Does the MLP stem do the same thing?}
The same diagnostic applies to \textsc{mlp}'s first-layer weight
$W^{(1)} \in \mathbb{R}^{d_{\text{hidden}} \times C}$: its columns are
the per-channel input directions before the GELU nonlinearity, and
their pairwise cosines measure how much the encoder separates
channels at the linear-stem stage versus deferring that separation
to GELU plus $W^{(2)}$. Training \textsc{mlp} at $C \in \{4, 8, 16\}$
with $5$ seeds and applying the off-diagonal cosine diagnostic to
$W^{(1)}$'s columns:

\begin{table}[h]
\centering
\caption{\textsc{mlp} first-layer column geometry ($300$ epochs,
$d_{\text{model}}{=}64$). This is a $5$-seed sub-sweep run for the
gram diagnostic, so the NLL column is noisier than the $20$-seed
main-table estimate (\textsc{mlp} at $C{=}4$ in Table~\ref{tab:main}
is $2.177 \pm 0.022$); the two estimates are within each other's
seed std. Compare to \textsc{linear}'s $W_k$ at $C{=}4$:
mean $0.042$, max $0.095$.}
\label{tab:mlp-gram}
\begin{tabular}{lccc}
\toprule
$C$  & best val NLL & mean $|\cos|$ & max $|\cos|$ \\
\midrule
$4$  & $2.170 \pm 0.036$ & $0.075 \pm 0.021$ & $0.158 \pm 0.058$ \\
$8$  & $2.214 \pm 0.016$ & $0.074 \pm 0.015$ & $0.248 \pm 0.098$ \\
$16$ & $2.222 \pm 0.025$ & $0.105 \pm 0.005$ & $0.463 \pm 0.109$ \\
\bottomrule
\end{tabular}
\end{table}

\textsc{mlp}'s linear stem is also pushed toward near-orthogonality
by the task loss, but consistently looser than the linear family's
$W_k$: at $C{=}4$ its mean off-diagonal cosine is $1.8\times$
\textsc{linear}'s ($0.075$ vs.\ $0.042$); at $C{=}16$ the gap
widens further and pairs of columns reach $|\cos| \approx 0.46$.
The mechanism is straightforward: \textsc{mlp} can also separate
channels through the GELU plus $W^{(2)}$, so the gradient pressure
on $W^{(1)}$ alone is partially relieved. The qualitative finding
survives---spontaneous channel orthogonalisation is a property of
the task loss interacting with a per-channel input stem, not
unique to a single-layer projection---but quantitatively the
linear family achieves the cleanest geometry.

\paragraph{Bias-ablation note.}
\label{sec:results-bias}
\textsc{linear}'s forward pass $\mathbf{h}(t) = \sum_k (W_k v_k(t)
+ b_k) + \mathbf{p}(t)$ has a redundancy: the $C \cdot
d_{\text{model}}$ per-channel biases sum to a single constant
offset that adds to every embedding regardless of input or
position, so they cannot affect the cosine geometry. Retraining
\textsc{linear} with the channel bias zeroed and frozen
(\textsc{linear-nobias}) confirms this experimentally: val NLL
stays within seed std ($2.177 \pm 0.013$ for \textsc{linear-nobias}
vs.\ $2.169 \pm 0.014$ for \textsc{linear}, both at $5$ seeds), and
mean/max $|\cos|$ match to three decimals. The \textsc{linear}
number here is a $5$-seed sub-sweep run for the ablation and is
slightly higher than the $20$-seed main-table estimate
($2.155 \pm 0.019$; both means within the same seed std). We
retain the bias in \textsc{linear}'s headline definition to match
the PyTorch \texttt{nn.Linear} default; the point of the ablation
is that the geometric argument is bias-independent.

\subsection{Encoding space allocation}
\label{sec:results-space}

Beyond separating channels, does the model allocate more encoding
space to channels that matter more? Using the norm and
variance-fraction diagnostics from
Section~\ref{sec:method-diagnostics}. At $C{=}4$ every channel is a
driver, so the driver/distractor split is trivial; we report
$C \in \{8, 16\}$ where the question becomes meaningful.

\begin{table}[h]
\centering
\caption{Encoding space allocation for \textsc{linear} ($300$
epochs, mean over $20$ seeds). Channels $0$--$3$ always drive the
outcome; the remaining $(C{-}4)$ channels are distractors.}
\label{tab:space}
\begin{tabular}{lcccc}
\toprule
& \multicolumn{2}{c}{$\|W_k\|$ norm} & \multicolumn{2}{c}{variance fraction} \\
\cmidrule(lr){2-3} \cmidrule(lr){4-5}
& drivers (0--3) & distractors & drivers & distractors \\
\midrule
$C{=}8$  & $1.22$ & $0.54$ & $83.8\%$ & $16.2\%$ \\
$C{=}16$ & $1.30$ & $0.59$ & $62.4\%$ & $37.6\%$ \\
\bottomrule
\end{tabular}
\end{table}

Driver channels receive ${\sim}2.2{\times}$ larger seed-averaged norms
than distractors at $C{=}8$ and ${\sim}2.1{\times}$ at $C{=}16$. The
four drivers capture $83\%$ of the embedding variance at $C{=}8$ and
$62\%$ at $C{=}16$, where the aggregate share is naturally diluted by
having twelve distractors but the per-channel variance contribution
remains markedly larger for drivers ($\sim$$15\%$ each vs.\
$\sim$$3\%$ each). We do not report per-channel error bars on the
norms, and the fine-grained ordering within drivers is descriptive
only; the robust claim is the driver/distractor gap, which holds at
both channel counts.

\paragraph{Why distractor norms are not zero.}
The gap is large but the distractor norms (${\sim}0.55$ at
$C{=}8$) do not vanish. This is a finite-data equilibrium: the
expected gradient on a distractor weight factorises to zero by
the distractor's independence from the outcome, but stochastic
estimates have $\mathcal{O}(1/\sqrt{N_{\text{series}}})$ variance,
and under L2 weight decay distractor weights sit at the
corresponding noise floor rather than being driven to zero. The
$1/\sqrt{N}$ scaling predicts a ${\sim}\sqrt{10} \approx 3.2$
reduction in distractor norms at $N{=}5120$ (the data-rich
condition of Section~\ref{sec:results-largen}); the observed
reduction is ${\sim}2.5\times$, close to but slightly slower than
predicted (driver/distractor norm ratio rises from $2.2\times$
to $6.7\times$, driver variance share from $83\%$ to $98\%$).

\subsection{Channel identifiability via linear probing}
\label{sec:results-probe}

\begin{table}[t]
\centering
\caption{Linear-probe channel recovery at $C{=}4$, seed 0, 300 epochs.
Closed-form ridge probe $\mathbf{h} \mapsto \hat{x}$ from a frozen hidden
state to the raw input $x \in \mathbb{R}^4$; held-out per-channel
$R^2$. Layer~0 is the input embedding; layer~3 is the final transformer
block. All per-channel-$W_k$ encoders achieve near-perfect input
recovery and preserve mean $R^2 \ge 0.84$ through three attention
layers; \textsc{sum} collapses to $R^2 \approx 0.24$ at every depth.}
\label{tab:probe}
\begin{tabular}{llccccc}
\toprule
Encoder & Layer & $R^2_{\text{ch0}}$ & $R^2_{\text{ch1}}$ & $R^2_{\text{ch2}}$ & $R^2_{\text{ch3}}$ & mean \\
\midrule
\textsc{sum}        & 0 (input) & $0.255$ & $0.250$ & $0.242$ & $0.214$ & $0.240$ \\
\textsc{sum}        & 3 (final) & $0.248$ & $0.242$ & $0.235$ & $0.206$ & $0.233$ \\
\textsc{linear}     & 0 (input) & $1.000$ & $1.000$ & $1.000$ & $1.000$ & $1.000$ \\
\textsc{linear}     & 3 (final) & $0.957$ & $0.861$ & $0.911$ & $0.879$ & $0.902$ \\
\textsc{linear-ortho}  & 0 (input) & $1.000$ & $1.000$ & $1.000$ & $1.000$ & $1.000$ \\
\textsc{linear-ortho}  & 3 (final) & $0.959$ & $0.856$ & $0.915$ & $0.885$ & $0.904$ \\
\textsc{mlp}        & 0 (input) & $1.000$ & $1.000$ & $1.000$ & $1.000$ & $1.000$ \\
\textsc{mlp}        & 3 (final) & $0.934$ & $0.879$ & $0.913$ & $0.646$ & $0.843$ \\
\textsc{linear-ppe} & 0 (input) & $1.000$ & $1.000$ & $1.000$ & $1.000$ & $1.000$ \\
\textsc{linear-ppe} & 3 (final) & $0.930$ & $0.861$ & $0.874$ & $0.847$ & $0.878$ \\
\textsc{concat}     & 0 (input) & $0.998$ & $1.000$ & $1.000$ & $1.000$ & $0.999$ \\
\textsc{concat}     & 3 (final) & $0.955$ & $0.852$ & $0.906$ & $0.845$ & $0.890$ \\
\bottomrule
\end{tabular}
\end{table}

Following the procedure in Section~\ref{sec:method-diagnostics},
Table~\ref{tab:probe} reports per-channel $R^2$ at layer~$0$ (the
encoder output) and layer~$3$ (the output of the final transformer
block).

First, every per-channel-$W_k$ encoder recovers the raw inputs
near-perfectly at layer~$0$: \textsc{linear}, \textsc{linear-ortho},
\textsc{linear-ppe}, and \textsc{mlp} all hit $R^2{=}1.000$
across all four channels, and \textsc{concat} does the same up to
rounding ($R^2_{\text{ch0}}{=}0.998$). The encoder is writing the
channels into a linearly recoverable subspace by construction.

Second, deep recoverability survives the full three-block stack:
mean $R^2$ at layer~$3$ stays at $0.84$--$0.90$ for every
per-channel-$W_k$ encoder. \textsc{mlp} drops to $0.646$ on ch3
while keeping the other three above $0.87$; the other encoders
degrade roughly uniformly. The result confirms the residual-stream
prediction in Section~\ref{sec:method-diagnostics}: the linear
channel directions the encoder writes at layer~$0$ persist through
every block.

Third, \textsc{sum} collapses to $R^2 \approx 0.24$ at every
depth. This is essentially the information-theoretic floor
$1/C = 0.25$ for recovering one of four independent unit-variance
channels from their pointwise sum: the information loss happens
at the encoder and is unrecoverable downstream.

\subsection{Channel masking}
\label{sec:results-mask}

\begin{table}[t]
\centering
\caption{Test-time channel ablation at $C{=}4$, seed 0, 300 epochs. We
zero a single channel's input at inference; rows show the resulting val
accuracy. $\Delta$ is accuracy drop from the unmasked baseline. The
outcome formula uses ch0 in two interaction terms, ch1 in one, and
ch2/ch3 once each with smaller marginal effect; the per-channel-$W_k$
encoders reproduce this ordering cleanly, whereas \textsc{sum},
\textsc{ci}, and \textsc{cat} are flatter.}
\label{tab:mask}
\begin{tabular}{lccccc}
\toprule
encoder & no mask & mask ch0 & mask ch1 & mask ch2 & mask ch3 \\
\midrule
\textsc{sum}        & $0.093$ & $0.082$ & $0.089$ & $0.084$ & $0.094$ \\
$\Delta$ acc        &   ---   & $-0.011$ & $-0.004$ & $-0.009$ & $+0.001$ \\
\textsc{ci}         & $0.144$ & $0.082$ & $0.108$ & $0.086$ & $0.134$ \\
$\Delta$ acc        &   ---   & $-0.062$ & $-0.036$ & $-0.058$ & $-0.010$ \\
\textsc{cat}        & $0.217$ & $0.095$ & $0.110$ & $0.110$ & $0.200$ \\
$\Delta$ acc        &   ---   & $-0.122$ & $-0.107$ & $-0.107$ & $-0.017$ \\
\textsc{concat}     & $0.241$ & $0.096$ & $0.113$ & $0.103$ & $0.202$ \\
$\Delta$ acc        &   ---   & $-0.145$ & $-0.128$ & $-0.138$ & $-0.039$ \\
\textsc{linear}     & $0.238$ & $0.096$ & $0.116$ & $0.106$ & $0.204$ \\
$\Delta$ acc        &   ---   & $-0.142$ & $-0.122$ & $-0.132$ & $-0.034$ \\
\textsc{linear-ortho}  & $0.239$ & $0.096$ & $0.117$ & $0.104$ & $0.205$ \\
$\Delta$ acc        &   ---   & $-0.143$ & $-0.122$ & $-0.135$ & $-0.034$ \\
\textsc{mlp}        & $0.240$ & $0.098$ & $0.111$ & $0.106$ & $0.209$ \\
$\Delta$ acc        &   ---   & $-0.142$ & $-0.129$ & $-0.134$ & $-0.031$ \\
\textsc{linear-ppe} & $0.256$ & $0.096$ & $0.115$ & $0.106$ & $0.216$ \\
$\Delta$ acc        &   ---   & $-0.160$ & $-0.141$ & $-0.150$ & $-0.040$ \\
\bottomrule
\end{tabular}
\end{table}

The outcome formula gives us a known importance ordering: ch0
appears in two terms, interacting with ch1 and gated by ch3; ch1
appears once (interacting with ch0); ch2 enters alone as a sine;
ch3 enters as a sign indicator on ch0. So we know a priori which
channels the model \emph{should} rely on most.
Zeroing one channel at test time gives us a behavioural test of
whether the encoder reproduces this ordering. Table~\ref{tab:mask}
shows that the top-tier encoders produce sharp, interpretable
per-channel accuracy drops that match the formula's importance
ordering. \textsc{sum} is nearly flat, as expected from the
information-theoretic argument: it has no recoverable channel
identity for the masking to disturb.

\subsection{Convergence and wall-clock cost}
\label{sec:results-convergence}

Two complementary worries about more involved encoders: (i)~that
they cost extra training time even when they don't change the
ceiling, and (ii)~the converse---that they might, despite tying or
losing on best-NLL, at least \emph{reach} that ceiling faster and
therefore be the right choice when compute or epochs are
constrained. From the 300-epoch traces (val NLL recorded every 20
epochs, $C{=}4$, 5 seeds) we extract two diagnostics: the epoch at
which each run first reaches within $0.05$ NLL of its own best, and
the wall-clock seconds per epoch, normalised to \textsc{linear}.

\begin{table}[h]
\centering
\caption{Convergence speed and wall-clock cost ($C{=}4$, 300 epochs,
5 seeds, single GPU). Epochs-to-target is rounded to the trace
resolution of 20 epochs. Cost is per-epoch wall time normalised to
\textsc{linear}.}
\label{tab:convergence}
\begin{tabular}{lccc}
\toprule
encoder & best NLL & epoch to NLL $+0.05$ & relative cost \\
\midrule
\textsc{linear}     & $2.170$ & $\sim$152 & $1.00\times$ \\
\textsc{linear-ortho}  & $2.170$ & $\sim$156 & $1.04\times$ \\
\textsc{mlp}        & $2.171$ & $\sim$144 & $1.01\times$ \\
\textsc{concat}     & $2.183$ & $\sim$152 & $1.00\times$ \\
\textsc{linear-ppe} & $2.116$ & $\sim$148 & $0.99\times$ \\
\midrule
\textsc{sum}        & $3.252$ & $\sim$20 (plateau) & $1.00\times$ \\
\textsc{ci}         & $3.054$ & $\sim$20 (plateau) & $2.3\times$ \\
\textsc{cat}        & $2.360$ & $\sim$180 & $5.2\times$ \\
\bottomrule
\end{tabular}
\end{table}

The five top-tier encoders reach within $0.05$ NLL of their final
value at the same trace point ($\sim$150 epochs) and cost
essentially the same per-epoch wall time---the transformer backbone
dominates, so encoder choice within this family is free. Neither
worry materialises: the more involved encoders do not slow
convergence (the MLP stem's $9{\times}$ extra parameters cost
nothing per epoch and reach target on the same trace point), but
they also do not converge faster than \textsc{linear}. There is no
``faster-to-target'' argument available for picking
\textsc{mlp}, \textsc{linear-ortho}, or \textsc{linear-ppe} over
\textsc{linear} within the top tier.

\textsc{ci} and \textsc{cat} pay substantial wall-clock costs from
their architectures, not their input encoders: ${\sim}2.3\times$
\textsc{linear} at $C{=}4$ for \textsc{ci} (scaling to
${\sim}8\times$ at $C{=}16$) and ${\sim}5\times$ at $C{=}4$ for
\textsc{cat} (scaling with $C^2 T^2$ to ${\sim}17\times$ at
$C{=}8$). \textsc{cat} also takes longer
(${\sim}180$ epochs vs.\ ${\sim}150$) to reach its
own---worse---best NLL on the synthetic benchmark.

\subsection{Positional projection: the orthogonalisation mechanism}
\label{sec:results-pospro}

\textsc{linear-ppe} achieves the lowest val NLL at $C{=}4$ and
$C{=}8$ and ties \textsc{mlp} at $C{=}16$ (the two are within each
other's seed std at $C{=}16$, where \textsc{mlp} has the lower
mean $2.231$ vs.\ \textsc{linear-ppe}'s $2.252$). Within the
linear family, \textsc{linear-ppe} edges \textsc{linear} at every
$C$. The paired-seed design (Section~\ref{sec:method-training})
gives one $\Delta_s$ per seed; at $20$ seeds the
\textsc{linear-ppe} vs.\ \textsc{linear} picture is:

\begin{itemize}
\item $C{=}4$: $\Delta = 0.041$, paired $t{=}6.60$,
  $p{=}2.6{\times}10^{-6}$, $95\%$ bootstrap CI $[+0.029, +0.053]$,
  19/20 favour ppe.
\item $C{=}8$: $\Delta = 0.026$, paired $p{=}3.7{\times}10^{-4}$,
  $95\%$ bootstrap CI $[+0.014, +0.038]$, 16/20.
\item $C{=}16$: $\Delta = 0.016$, paired $p{=}0.036$, $95\%$ bootstrap
  CI $[+0.002, +0.028]$, 14/20.
\end{itemize}

The gap shrinks with $C$ but does not vanish; even at $C{=}16$, where
the earlier 5-seed analysis read as null, the 20-seed paired test
puts $\Delta$ above zero, just barely. The advantage is small in
magnitude---$\sim$$2\%$ of baseline NLL at $C{=}4$, $\sim$$0.7\%$ at
$C{=}16$---and the practical-significance reading is more subdued
than the statistical-significance one would suggest in isolation.

\paragraph{Direct measurement: orthogonalisation, not compression.}
There are two natural mechanisms by which a learned linear
projection on top of $\mathbf{p}(t)$ could help. The
\emph{orthogonalisation} reading: $W_{\text{pos}}$ rotates the
positional basis out of $\mathrm{span}(W)$, so position and
channels stop competing for the same coordinates of the residual
stream. The \emph{compression} reading: $W_{\text{pos}}$ projects
$\mathbf{p}(t)$ onto a small effective subspace, leaving the
remaining dimensions free for channels. The two predictions differ:
orthogonalisation should reduce the overlap between the
positional basis and $\mathrm{span}(W)$ while leaving the basis's
intrinsic rank essentially unchanged; compression should reduce
that rank. We test both directly.

We extract the per-channel projection matrix $W \in \mathbb{R}^{C
\times d_{\text{model}}}$ and the effective positional basis
$P \in \mathbb{R}^{T \times d_{\text{model}}}$ from trained
\textsc{linear} and \textsc{linear-ppe} models ($C{=}4$, 20 seeds
each). For \textsc{linear}, $P_{t,:} = \mathbf{p}(t)$ is the fixed
sinusoidal basis; for \textsc{linear-ppe}, $P_{t,:} =
W_{\text{pos}} \mathbf{p}(t) + \mathbf{b}_{\text{pos}}$ is the
learned-rotated basis. Two metrics suffice to discriminate the
mechanism (Table~\ref{tab:pospro-geom}):

\begin{table}[h]
\centering
\caption{Geometric measurement of the positional basis $P$ and its
overlap with the channel subspace $\mathrm{span}(W)$
($C{=}4$, $300$ epochs, mean over 20 seeds with $95\%$ bootstrap
CIs in brackets where seed variance is non-zero).}
\label{tab:pospro-geom}
\begin{tabular}{lcc}
\toprule
& \textsc{linear} & \textsc{linear-ppe} \\
\midrule
Effective rank of $P$ (entropy of $\tilde\sigma^2$) & $7.59$ & $9.55$ $[9.17, 9.97]$ \\
Fraction of $\|P\|_F^2$ within $\mathrm{span}(W)$   & $0.034$ $[0.029, 0.039]$ & $0.005$ $[0.005, 0.006]$ \\
\bottomrule
\end{tabular}
\end{table}

The two metrics give opposite verdicts on the two readings.
Orthogonalisation is confirmed: the fraction of $P$'s energy that
lies inside $\mathrm{span}(W)$ drops by $\sim$$6.3\times$ under
\textsc{linear-ppe} ($3.4\% \to 0.5\%$; paired-difference $95\%$
bootstrap CI on the reduction $[-0.034, -0.024]$), so the learned
$W_{\text{pos}}$ rotation actively pushes the positional encoding
out of the channel subspace. Compression is contradicted: $P$'s
effective rank is \emph{higher} under \textsc{linear-ppe} ($9.55$
vs.\ $7.59$), not lower---the rotation spreads positional energy
across more directions, not fewer. The mechanism is rotation, not
compression.

The redistribution is visible at the level of individual singular
values: \textsc{linear}'s $P$ has one dominant direction ($\sigma_0
\approx 52$) plus a long tail; \textsc{linear-ppe}'s is more evenly
distributed ($\sigma_0 \approx 25$, with the next several around
$11$--$13$). Rotating the dominant positional mode out of
$\mathrm{span}(W)$ inevitably redistributes its energy across other
directions, which is the same effect read either as ``orthogonalising
position from channels'' or as ``flattening the singular spectrum of
$P$''.

(Principal angles between $\mathrm{span}(W)$ and $\mathrm{span}(P)$
collapse to $\sim$$0°$ at this dimensionality---$\mathrm{span}(W)$
is $C$-dimensional and trivially contained in
$\mathrm{span}(P)$'s $\sim$$22$-rank subspace---so they measure
subset membership rather than energetic overlap and are not the
right tool here.)

With orthogonalisation confirmed, the $C$-dependent shrinkage of
the gap has a clean mechanistic reading: $W_{\text{pos}}$ must
rotate $P$ out of a $C$-dimensional channel span using a fixed
$d_{\text{model}}^2$ parameter budget, and the rotation becomes
increasingly over-constrained as $C$ grows. At $C{=}4$ there are
$d_{\text{model}} - C = 60$ ``free'' dimensions for $P$ to occupy;
at $C{=}16$ only $48$, with the channel subspace also pressing
harder on the residual stream overall. The same mechanism explains
why the lead grows mildly with $d_{\text{model}}$ in the
Section~\ref{sec:results-dmodel} sweep---more residual-stream
capacity gives the rotation more directions to move into. Two
further factors---only four channels are drivers regardless of
$C$, and seed-level optimisation variance grows with $C$---could
contribute, but neither directly invokes the rotation mechanism
this section establishes.

\subsection{Loss-family robustness check: MSE target}
\label{sec:results-mse}

\begin{table}[t]
\centering
\caption{Loss-family robustness check: MSE/regression target replaces
categorical cross-entropy. Same backbone, $5$ seeds, $300$ epochs,
$d_{\text{model}}{=}64$. Lower MSE is better. The ordering survives
the swap: \textsc{sum} collapses ($R^2 \approx 0.13$ at $C{=}4$,
$\approx 0.01$ at $C{=}16$), the per-channel-$W_k$ tier
(\textsc{concat}, \textsc{linear}, \textsc{linear-ortho}, \textsc{mlp},
\textsc{linear-ppe}) clusters tight, and \textsc{linear-ppe} edges
\textsc{linear} at $C{=}4$ ($\sim$$6\%$ MSE reduction). \textsc{mlp}
pulls ahead of \textsc{linear} at $C{=}16$, mirroring the categorical
sweep.}
\label{tab:mse}
\begin{tabular}{lcccc}
\toprule
  & \multicolumn{2}{c}{$C{=}4$} & \multicolumn{2}{c}{$C{=}16$} \\
\cmidrule(lr){2-3}\cmidrule(lr){4-5}
encoder & val MSE $\downarrow$ & val $R^2$ $\uparrow$ & val MSE $\downarrow$ & val $R^2$ $\uparrow$ \\
\midrule
\textsc{sum}              & $0.4719 \pm 0.0090$ & $0.128 \pm 0.016$ & $0.5338 \pm 0.0167$ & $0.011 \pm 0.006$ \\
\textsc{concat}           & $0.0636 \pm 0.0027$ & $0.882 \pm 0.006$ & $0.0862 \pm 0.0071$ & $0.840 \pm 0.014$ \\
\textsc{linear}           & $0.0619 \pm 0.0024$ & $0.886 \pm 0.005$ & $0.0736 \pm 0.0026$ & $0.863 \pm 0.008$ \\
\textsc{linear-ortho}     & $0.0619 \pm 0.0024$ & $0.886 \pm 0.005$ & $0.0735 \pm 0.0020$ & $0.864 \pm 0.007$ \\
\textsc{mlp}              & $0.0643 \pm 0.0044$ & $0.881 \pm 0.009$ & $0.0678 \pm 0.0025$ & $0.874 \pm 0.007$ \\
\textsc{linear-ppe}       & $0.0582 \pm 0.0031$ & $0.892 \pm 0.007$ & $0.0722 \pm 0.0033$ & $0.866 \pm 0.007$ \\
\bottomrule
\end{tabular}
\end{table}

The categorical $K{=}32$-bin head could in principle interact with
encoder choice---e.g.\ by rewarding encoders that resolve fine
target quantiles---so we rerun a subset of the synthetic main sweep
with a scalar regression head trained on the continuous
(pre-binning) target under MSE loss. Same backbone, same training
schedule, $5$ seeds, $C \in \{4, 16\}$, and the same encoder set
as the $d_{\text{model}}$ sweep (\textsc{sum}, \textsc{concat},
\textsc{linear}, \textsc{linear-ortho}, \textsc{mlp},
\textsc{linear-ppe}); see Table~\ref{tab:mse}. The ordering is
preserved at both channel counts: \textsc{sum} collapses to $R^2
\approx 0.13$ at $C{=}4$ and $\approx 0.01$ at $C{=}16$, consistent
with the information-theoretic ceiling argument; the
per-channel-$W_k$ tier clusters tight; \textsc{linear-ppe} edges
\textsc{linear} at $C{=}4$ ($\sim$$6\%$ relative reduction in MSE,
$0.058$ vs.\ $0.062$), and \textsc{mlp} edges \textsc{linear} at
$C{=}16$. The
categorical-vs.-MSE choice is therefore not what is driving the
encoder ordering---both losses penalise \textsc{sum} for the same
information-theoretic reason, and both reveal the same small gaps
inside the top tier. The MSE sweep is not powered (5 seeds, no
paired design) to resolve the within-tier gaps at the statistical
level of the categorical 20-seed analysis; what it does is rule out
that the entire ordering is a label-binning artefact.

\subsection{Real-data validation: ETTh1}
\label{sec:results-etth1}

\begin{table}[t]
\centering
\caption{ETTh1 next-step bin prediction. 7 variates, $T{=}160$, $K{=}32$
quantile bins on the target OT (oil temperature) computed from the train
split. Target is the next-step bin of OT; all 7 variates are inputs.
$d_{\text{model}}{=}56$, 7 heads, 3 layers, $d_{\text{ff}}{=}224$; 300
epochs, cosine decay, mean $\pm$ std over 20 seeds (5 canonical + 15
paired-seed extras).}
\label{tab:etth1}
\begin{tabular}{lcc}
\toprule
encoder & val NLL $\downarrow$ & val acc $\uparrow$ \\
\midrule
\textsc{sum}        & $3.668 \pm 0.063$  & $0.016 \pm 0.020$ \\
\textsc{ci}         & $0.865 \pm 0.038$  & $0.664 \pm 0.018$ \\
\midrule
\textsc{cat}        & $0.551 \pm 0.019$  & $0.785 \pm 0.010$ \\
\textsc{linear}     & $0.561 \pm 0.017$  & $0.786 \pm 0.008$ \\
\textsc{linear-ortho}  & $0.561 \pm 0.017$  & $0.784 \pm 0.009$ \\
\textsc{concat}     & $0.571 \pm 0.019$  & $0.783 \pm 0.013$ \\
\textsc{linear-ppe} & $0.573 \pm 0.014$  & $0.776 \pm 0.009$ \\
\textsc{mlp}        & $0.585 \pm 0.020$  & $0.788 \pm 0.011$ \\
\bottomrule
\end{tabular}
\end{table}

Table~\ref{tab:etth1} mostly confirms the synthetic ordering on
ETTh1: the per-channel-$W_k$ family (\textsc{linear},
\textsc{linear-ortho}, \textsc{concat}, \textsc{linear-ppe},
\textsc{mlp}) cluster within one standard deviation across seeds of
one another, \textsc{ci} underperforms, and \textsc{sum} fails
catastrophically (NLL near $\ln 32 \approx 3.47$).

One difference from the synthetic ranking: \textsc{cat}, which sits
in the middle of the synthetic table, here posts the lowest mean NLL
($0.551$). Paired-difference tests at 20 seeds give a nuanced
picture:

\begin{itemize}
\item Decisive: \textsc{cat} beats \textsc{mlp}
  (paired $p{=}1.4{\times}10^{-7}$, 20/20) and \textsc{linear-ppe}
  ($p{=}3{\times}10^{-4}$, 17/20).
\item Marginal: \textsc{cat} beats \textsc{concat}
  ($p{=}0.002$, 16/20).
\item Indistinguishable: \textsc{cat} vs.\ \textsc{linear}
  ($\Delta = 0.010$, paired $p{=}0.14$, 13/20) and
  \textsc{cat} vs.\ \textsc{linear-ortho}
  ($\Delta = 0.010$, paired $p{=}0.10$, 13/20).
\end{itemize}

Best-bin accuracies across the top tier are indistinguishable
($0.78$--$0.79$ with standard deviation across seeds ${\sim}0.01$).
So \textsc{cat} posts
the lowest mean NLL but is statistically tied with \textsc{linear}
and \textsc{linear-ortho} at the seed level we test. The finding
is not that channel-as-token decisively wins on real data; rather,
its synthetic-benchmark disadvantage closes and a small
directional lead emerges that does not survive a paired test
against the closest competitors. A plausible reading is that ETTh1's seven variates are all
informative measurements of the same underlying electrical system
(oil temperature plus six grid-load variables)---unlike the
synthetic benchmark, which deliberately mixes drivers with
independent distractors---so the fine-grained cross-variate
attention \textsc{cat} enables has more to work with. Whether that pays \emph{enough} to prefer it over the
per-channel-$W_k$ default at matched wall-clock budget is much
less clear.

\section{Discussion}

\paragraph{\textsc{Linear} is not new.}
Stacking per-channel projections $W_k$ and summing is
\texttt{nn.Linear(C, $d_{\text{model}}$)}---the most obvious default.
Our contribution is not proposing it but auditing it: at matched
parameter budget, no variant we test substantially dislodges it.
On the synthetic benchmark, \textsc{concat} and \textsc{linear-ortho}
tie it within seed noise, \textsc{linear-ppe} edges it by
${\sim}2\%$ NLL at every $C$ (Section~\ref{sec:results-scaling}),
\textsc{mlp} matches it at $C{=}4$ and edges it narrowly at $C{=}16$,
and the architectural alternatives \textsc{ci} and \textsc{cat}
underperform decisively. On ETTh1 it ties at the top of the
per-channel-$W_k$ tier. Only the shared-projection \textsc{sum}
consistently loses outright.

\paragraph{What fails about the shared-scalar baseline.}
\textsc{sum} collapses because, with a shared scalar projection $W$
and constant per-channel embeddings, the encoder depends on its
inputs only through their pointwise sum $S(t) = \sum_k v_k(t)$.
That map has a $(C{-}1)$-dimensional null space, and by the
data-processing inequality no downstream layer can recover what is
destroyed before the transformer sees it. The probe makes the
floor explicit: for independent unit-variance channels at $C{=}4$
the theoretical ceiling for linearly recovering one channel from
$S$ is $\rho^2 = 1/C = 0.25$, and the layer-0 probe of
\textsc{sum} sits at $R^2 \approx 0.24$, touching it
(Table~\ref{tab:probe}). The $15{\times}$ $d_{\text{model}}$
sweep in Table~\ref{tab:dmodel} moves NLL by less than $0.005$
because the extra parameters never get to see the lost information.
The bottleneck is not summation per se---\textsc{linear} also
sums---but the \emph{shared} projection, which collapses the
per-channel directions into one before the sum.

\paragraph{The residual stream carries channel identity end-to-end.}
Linear probes at layer 3 recover each channel to $R^2 \ge 0.84$,
confirming the residual-stream prediction in
Section~\ref{sec:method-diagnostics}: per-channel-$W_k$ encoders
write near-orthogonal channel directions into the residual stream
at layer~$0$ and the pre-LN architecture preserves them through
every block essentially by construction, since LayerNorm only
rescales and the additive residual keeps them in the basis. This
is the mechanistic-interpretability picture of the residual stream
as an additive communication channel
\citep{elhage2021framework} applied at the level of input
channels: the encoder chooses the subspace, the backbone passes
it forward, and the deep-probe recoverability falls out of both.

\paragraph{Reading for non-numerical inputs.}
The mechanism is geometric and encoding-agnostic, which suggests
one natural extension worth flagging: when the per-channel inputs
are precomputed embeddings from an outside source (e.g.\ text,
image, or tabular feature embeddings fed into a multi-stream
transformer), the \textsc{linear-ppe}-style learned rotation has a
ready analog as a per-stream rotation that pushes precomputed
embedding subspaces off the positional subspace and off one
another. Whether this gives gains comparable to the
$\sim$$2\%$ NLL improvement we measure here, or larger because
external embeddings carry more structured competing geometry, is
an interesting question for another study.

\paragraph{Statistical significance vs.\ practical
near-equivalence.}
The $20$-seed paired tests reported above are powerful enough to
resolve several small inter-encoder gaps; it is worth separating
that statistical power from a practical claim. Among the
per-channel-$W_k$ family (\textsc{linear}, \textsc{linear-ortho},
\textsc{concat}, \textsc{mlp}, \textsc{linear-ppe}) at $C{=}4$ the
absolute NLL differences span $0.02$ on a baseline of $2.16$;
\textsc{linear-ppe}'s lead over \textsc{linear} is $0.041$ NLL
($\sim$$2\%$ of baseline); \textsc{mlp}'s lead over the linear
family at $C{=}16$ is $0.010$--$0.023$ in the data-rich regime.
These are real gaps under our experimental design, but they sit
inside the spread that typical task-design choices (channel
count, sequence length, label binning, etc.) would induce, and we
would not expect them to translate cleanly to practically
significant differences on new tasks with new data distributions.

\section{Limitations}

\begin{itemize}
\item The synthetic benchmark deliberately makes channel identity
      informative. On tasks with exchangeable channels the encoder
      design problem itself changes---channel identity ceases to
      be a useful signal---so \textsc{sum} is not penalised and
      the per-channel-$W_k$ family's advantage over it should
      shrink or disappear entirely.
\item Main sweep at $d_{\text{model}}{=}64$, $C \in \{4,8,16\}$;
      ETTh1 at $d_{\text{model}}{=}56$, $C{=}7$. The $d_{\text{model}}$
      sweep covers all top-tier encoders plus \textsc{sum} at
      $d \in \{64, 128, 256\}, C{=}4$, but all encoders overfit at
      $d \ge 128$ under the main-sweep training data
      (Table~\ref{tab:dmodel}); the reported best-NLL numbers there
      are effective-early-stopping minima rather than converged
      values.
\item The main sweep at $N_{\text{series}}{=}512$ is data-limited.
      $10\times$ training data improves all top-tier encoders by
      ${\sim}0.4$ NLL at $C{=}16$ (Table~\ref{tab:main-largen}).
      The relative ordering of encoders is robust to data scale,
      but absolute inter-encoder gaps shrink ${\sim}5\times$ at
      $10\times$ data; readers should treat the $N{=}512$
      inter-encoder gaps as upper bounds on the data-rich-regime
      gaps.
\item \textsc{cat} is skipped at $C{=}16$ on the synthetic benchmark
      for compute reasons ($\mathcal{O}((CT)^2)$ attention).
\item Headline comparisons are at $20$ paired seeds. A few
      diagnostics (linear probing, channel masking, channel-bias
      ablation, \textsc{mlp} first-layer geometry, MSE loss-family
      check) remain at $5$ seeds, labelled as such in their tables.
\end{itemize}

\section{Related work}

\paragraph{Channel-as-token and channel-independent TSF.}
iTransformer \citep{liu2024itransformer} treats each variate as a
token; PatchTST \citep{nie2023patchtst} runs independent backbones
per channel; Crossformer \citep{zhang2023crossformer} does both.
These preserve per-channel capacity at the cost of longer sequences
or per-channel compute. Our finding is that on the synthetic
benchmark a per-channel linear stem on a shared backbone (one
token per time step, $d_{\text{model}}$-dim) matches both
architectures at $C{=}4, 8, 16$; on ETTh1 \textsc{cat} posts the
lowest mean NLL but is statistically tied with \textsc{linear} and
\textsc{linear-ortho} under paired analysis.
\citet{han2024capacity} reach the opposite headline for channel
independence: comparing CI and channel-dependent training across
standard long-horizon benchmarks (including ETTh1), they find CI
consistently \emph{wins}, and attribute this to a
capacity--robustness trade-off---channel-dependent models have
higher capacity but are fragile under the distribution drift that
dominates those benchmarks. We read our \textsc{ci} result as the
capacity side of that same trade-off rather than a contradiction:
our tasks are next-step prediction in which cross-channel
structure is genuinely informative and stable, so the capacity
ceiling of channel-independent processing binds and \textsc{ci}
pays for it. Which side of the trade-off dominates is a property
of the data regime, not of the architecture alone.
\citet{qiu2025channelstrategy} survey the field's channel
strategies along exactly this independent/dependent/partial axis;
our encoder set operationalises several points on that axis inside
one controlled architecture.

\paragraph{Component-level benchmarking.}
TimeRecipe \citep{zhao2025timerecipe} benchmarks
forecasting-module choices---including input-embedding families
(token, patch, inverted, frequency)---across more than $10{,}000$
runs, asking which module combinations work in which regimes. Our
audit is narrower and deeper: we hold the backbone fixed and
isolate the per-time-step channel-embedding site, use seed-paired
statistics, and probe the geometric mechanism behind the observed
near-equivalence rather than ranking configurations.

\paragraph{Linear and MLP stems in time-series forecasting.}
Many multivariate transformers use a linear input stem
\citep{zhou2021informer,wu2023timesnet}, and a parallel line of
work argues that the transformer machinery is itself often
unnecessary: \citet{zeng2023transformers} show that simple linear
forecasters (\textsc{DLinear}, \textsc{NLinear}) match or beat
several transformer architectures on standard multivariate
benchmarks, and \citet{chen2023tsmixer} make the same point with
an all-MLP architecture (\textsc{TSMixer}) that separates
channel-mixing from time-mixing. Our work is on the orthogonal
question of how the input layer of a transformer should embed
multiple channels rather than whether a transformer is needed at
all; the two lines of evidence are compatible---a linear input
projection is also a strong default---and our \textsc{linear}
baseline carries that lesson inside the encoder even when the
backbone is non-linear. The relationship to \textsc{TSMixer} is
particularly direct: its channel-mixing MLP is essentially our
\textsc{mlp} encoder used as the centrepiece rather than as an
input stem.

\paragraph{Orthogonality regularisation and emergent orthogonality.}
\citet{bansal2018gainsorthogonality} use soft-orthogonality
penalties for feature decorrelation in deep networks. We apply the
same idea to per-channel input projections and find it
unnecessary: the task loss produces the same geometry. That
near-orthogonal arrangements can emerge without a penalty has
precedent outside time series. In the superposition framework of
\citet{elhage2022superposition}, a model with at least as many
dimensions as features is expected to place features along
near-orthogonal directions---our $C \ll d_{\text{model}}$ setting
sits squarely in that regime---and \citet{guha2026capacity} report
that for language-model token embeddings, near-orthogonality is
largely inherited from random initialisation (a high-dimensional
default) rather than produced by training. This is why
Section~\ref{sec:results-gram} reports the at-initialisation
baseline explicitly: in our setting training tightens the
off-diagonal cosines well below the random-initialisation level,
so the effect is active orthogonalisation, not mere persistence
of the initial geometry.

\paragraph{Channel-importance weighting.}
Variable Selection Networks in the Temporal Fusion Transformer
\citep{lim2021tft} explicitly learn per-channel importance weights
via a gating mechanism on top of per-channel embeddings, on the
hypothesis that channels need an architectural mechanism to be
weighted by relevance. Our encoding-space-allocation result
(Section~\ref{sec:results-space}) is partly negative evidence for
that hypothesis in the synthetic setting: a bare
\texttt{nn.Linear} discovers the driver/distractor split through
its learned $\|W_k\|$ alone, with no explicit weighting
mechanism. We do not claim this generalises to the harder
settings TFT was designed for---static covariates, mixed
categorical/continuous inputs, long-horizon multi-step
forecasting---but at least for the homogeneous-numerical input
case the explicit selection mechanism is solving a problem the
encoder already solves. The implicit mechanism itself has
independent precedent: \citet{radhakrishnan2024mechanism} show
that trained first-layer weights track the average gradient outer
product of the learned function, up-weighting influential inputs
with no selection machinery; our norm-based allocation result
reads as the channel-encoder instance of that general effect.

\paragraph{Decoupling position from content.}
\citet{ke2021rethinking} showed that untying positional embeddings
from content embeddings---routing position through a separate
projection rather than additively combining them at the input---
improves transformer language models. \textsc{linear-ppe} applies
the same principle to the channel--position competition in
time-series transformers, and the direct geometric measurement in
Section~\ref{sec:results-pospro} identifies the mechanism as the
same one: position is rotated out of the channel subspace so the
two stop competing for the same residual-stream coordinates.
Related devices appear across domains. DeBERTa
\citep{he2021deberta} keeps content and position in separate
vectors with disentangled attention. Transformer-TTS
\citep{li2019transformertts} rescales the sinusoidal encoding by a
single trained scalar to match the embedding's scale---a
degenerate case of \textsc{linear-ppe}'s map, and a useful
contrast: a scalar can rescale the positional basis but cannot
rotate it out of the channel subspace, which
Section~\ref{sec:results-pospro} identifies as the operative
mechanism. Learnable Fourier features \citep{li2021learnable}
train both the frequencies and a nonlinear map on top of a
Fourier positional representation, of which \textsc{linear-ppe}
is the minimal fixed-frequency, linear member. For a survey of
positional-encoding variants specific to time-series
transformers, see \citet{irani2025positional}.

\section{Conclusion}

The encoder design problem turns out to be largely settled by the
task loss itself. A per-channel linear projection
(\texttt{nn.Linear(C, $d_{\text{model}}$)}) matches every more
elaborate encoder we tested up to small, statistically real but
practically modest, differences. The task loss drives the channel
projections to near-orthogonality and scales their norms with
channel importance; a shared-scalar projection has a
capacity-independent information-theoretic ceiling. The exceptions
resolved under paired analysis are small and qualified:
\textsc{linear-ppe} leads the linear family at every $C$ by
${\sim}2\%$ NLL via positional--channel orthogonalisation,
\textsc{mlp} edges them at $C{=}16$ with the gap shrinking
${\sim}3{\times}$ under $10{\times}$ training data, and
\textsc{cat} posts the lowest mean NLL on ETTh1 but is
statistically tied with \textsc{linear} and \textsc{linear-ortho}.

\textbf{Practical recommendation.} Default to
\texttt{nn.Linear(C, $d_{\text{model}}$)}: within the
per-channel-$W_k$ family the encoder choice is, for any practical
purpose, close to a free parameter, and the simplest option is the
path of least surprise. \textsc{linear-ppe} is a justified
opportunistic extra when $d_{\text{model}}^2$ parameters are cheap
relative to the backbone. Reach for \textsc{mlp} or \textsc{cat}
only when the task gives a real reason---a high-channel regime
where nonlinear gating may help, or a setting whose cross-variate
structure justifies \textsc{cat}'s ${\sim}5{\times}$ wall-clock
cost and $C{\cdot}T$ context length.

Recognising that a per channel linear projection is all that
is needed for serving the input signals for the main model as
orthogonally separated entities, we can focus
our modelling efforts and model complexity to where it matters.

\bibliographystyle{plainnat}

\begin{thebibliography}{9}
\bibitem[Alain and Bengio(2017)]{alain2017probes}
G.~Alain and Y.~Bengio.
\newblock Understanding intermediate layers using linear classifier
  probes.
\newblock In \emph{ICLR Workshop}, 2017.

\bibitem[Bansal et al.(2018)]{bansal2018gainsorthogonality}
N.~Bansal, X.~Chen, and Z.~Wang.
\newblock Can we gain more from orthogonality regularizations in training
  deep networks?
\newblock In \emph{NeurIPS}, 2018.

\bibitem[Chen et al.(2023)]{chen2023tsmixer}
S.-A. Chen, C.-L. Li, N.~Yoder, S.~O. Ar{\i}k, and T.~Pfister.
\newblock {TSMixer}: an all-{MLP} architecture for time series
  forecasting.
\newblock \emph{Transactions on Machine Learning Research}, 2023.

\bibitem[Elhage et al.(2021)]{elhage2021framework}
N.~Elhage, N.~Nanda, C.~Olsson, T.~Henighan, N.~Joseph, B.~Mann,
  A.~Askell, Y.~Bai, A.~Chen, T.~Conerly, N.~DasSarma,
  D.~Drain, D.~Ganguli, Z.~Hatfield-Dodds, D.~Hernandez,
  A.~Jones, J.~Kernion, L.~Lovitt, K.~Ndousse, D.~Amodei,
  T.~Brown, J.~Clark, J.~Kaplan, S.~McCandlish, and C.~Olah.
\newblock A mathematical framework for transformer circuits.
\newblock \emph{Transformer Circuits Thread}, 2021.

\bibitem[Elhage et al.(2022)]{elhage2022superposition}
N.~Elhage, T.~Hume, C.~Olsson, N.~Schiefer, T.~Henighan,
  S.~Kravec, Z.~Hatfield-Dodds, R.~Lasenby, D.~Drain, C.~Chen,
  R.~Grosse, S.~McCandlish, J.~Kaplan, D.~Amodei, M.~Wattenberg,
  and C.~Olah.
\newblock Toy models of superposition.
\newblock \emph{Transformer Circuits Thread}, 2022.

\bibitem[Guha(2026)]{guha2026capacity}
A.~Guha.
\newblock Representational capacity: geometric limits on feature
  representation in transformer language models.
\newblock arXiv:2606.02765, 2026.

\bibitem[Han et al.(2024)]{han2024capacity}
L.~Han, H.-J. Ye, and D.-C. Zhan.
\newblock The capacity and robustness trade-off: revisiting the
  channel independent strategy for multivariate time series
  forecasting.
\newblock \emph{IEEE Transactions on Knowledge and Data
  Engineering}, 2024.

\bibitem[He et al.(2021)]{he2021deberta}
P.~He, X.~Liu, J.~Gao, and W.~Chen.
\newblock {DeBERTa}: decoding-enhanced {BERT} with disentangled
  attention.
\newblock In \emph{ICLR}, 2021.

\bibitem[Irani and Metsis(2025)]{irani2025positional}
H.~Irani and V.~Metsis.
\newblock Positional encoding in transformer-based time series
  models: a survey.
\newblock arXiv:2502.12370, 2025.

\bibitem[Ke et al.(2021)]{ke2021rethinking}
G.~Ke, D.~He, and T.-Y. Liu.
\newblock Rethinking positional encoding in language pre-training.
\newblock In \emph{ICLR}, 2021.

\bibitem[Lim et al.(2021)]{lim2021tft}
B.~Lim, S.~\"O. Ar{\i}k, N.~Loeff, and T.~Pfister.
\newblock Temporal fusion transformers for interpretable multi-horizon
  time series forecasting.
\newblock \emph{International Journal of Forecasting},
  37(4):1748--1764, 2021.

\bibitem[Li et al.(2019)]{li2019transformertts}
N.~Li, S.~Liu, Y.~Liu, S.~Zhao, M.~Liu, and M.~Zhou.
\newblock Neural speech synthesis with transformer network.
\newblock In \emph{AAAI}, 2019.

\bibitem[Li et al.(2021)]{li2021learnable}
Y.~Li, S.~Si, G.~Li, C.-J. Hsieh, and S.~Bengio.
\newblock Learnable {Fourier} features for multi-dimensional
  spatial positional encoding.
\newblock In \emph{NeurIPS}, 2021.

\bibitem[Liu et al.(2024)]{liu2024itransformer}
Y.~Liu, T.~Hu, H.~Zhang, H.~Wu, S.~Wang, L.~Ma, and M.~Long.
\newblock {iTransformer}: inverted transformers are effective for time
  series forecasting.
\newblock In \emph{ICLR}, 2024.

\bibitem[Nie et al.(2023)]{nie2023patchtst}
Y.~Nie, N.~H.~Nguyen, P.~Sinthong, and J.~Kalagnanam.
\newblock A time series is worth 64 words: long-term forecasting with
  transformers.
\newblock In \emph{ICLR}, 2023.

\bibitem[Qiu et al.(2025)]{qiu2025channelstrategy}
X.~Qiu, H.~Cheng, X.~Wu, J.~Lu, J.~Hu, C.~Guo, C.~S. Jensen, and
  B.~Yang.
\newblock A comprehensive survey of deep learning for multivariate
  time series forecasting: a channel strategy perspective.
\newblock arXiv:2502.10721, 2025.

\bibitem[Radhakrishnan et al.(2024)]{radhakrishnan2024mechanism}
A.~Radhakrishnan, D.~Beaglehole, P.~Pandit, and M.~Belkin.
\newblock Mechanism for feature learning in neural networks and
  backpropagation-free machine learning models.
\newblock \emph{Science}, 383(6690):1461--1467, 2024.

\bibitem[Vaswani et al.(2017)]{vaswani2017attention}
A.~Vaswani, N.~Shazeer, N.~Parmar, J.~Uszkoreit, L.~Jones, A.~N. Gomez,
{\L}.~Kaiser, and I.~Polosukhin.
\newblock Attention is all you need.
\newblock In \emph{NeurIPS}, 2017.

\bibitem[Wu et al.(2023)]{wu2023timesnet}
H.~Wu, T.~Hu, Y.~Liu, H.~Zhou, J.~Wang, and M.~Long.
\newblock {TimesNet}: temporal 2{D}-variation modeling for general time
  series analysis.
\newblock In \emph{ICLR}, 2023.

\bibitem[Xiong et al.(2020)]{xiong2020layernorm}
R.~Xiong, Y.~Yang, D.~He, K.~Zheng, S.~Zheng, C.~Xing, H.~Zhang,
  Y.~Lan, L.~Wang, and T.-Y. Liu.
\newblock On layer normalization in the transformer architecture.
\newblock In \emph{ICML}, 2020.

\bibitem[Zeng et al.(2023)]{zeng2023transformers}
A.~Zeng, M.~Chen, L.~Zhang, and Q.~Xu.
\newblock Are transformers effective for time series forecasting?
\newblock In \emph{AAAI}, 2023.

\bibitem[Zhao et al.(2025)]{zhao2025timerecipe}
Z.~Zhao, J.~Ni, S.~Xu, H.~Liu, W.~Jin, and B.~A. Prakash.
\newblock {TimeRecipe}: a time-series forecasting recipe via
  benchmarking module level effectiveness.
\newblock arXiv:2506.06482, 2025.

\bibitem[Zhang and Yan(2023)]{zhang2023crossformer}
Y.~Zhang and J.~Yan.
\newblock {Crossformer}: transformer utilizing cross-dimension dependency
  for multivariate time series forecasting.
\newblock In \emph{ICLR}, 2023.

\bibitem[Zhou et al.(2021)]{zhou2021informer}
H.~Zhou, S.~Zhang, J.~Peng, S.~Zhang, J.~Li, H.~Xiong, and W.~Zhang.
\newblock {Informer}: beyond efficient transformer for long sequence
  time-series forecasting.
\newblock In \emph{AAAI}, 2021.
\end{thebibliography}

\end{document}